\documentclass[10pt,twocolumn,letterpaper]{article}

\usepackage[pagenumbers]{cvpr} % To force page numbers, e.g. for an arXiv version

\definecolor{cvprblue}{rgb}{0.21,0.49,0.74}
\usepackage[pagebackref,breaklinks,colorlinks,allcolors=cvprblue]{hyperref}
\usepackage{placeins}
\usepackage{multirow}
\usepackage{siunitx}
\usepackage[accsupp]{axessibility}
 
\newcommand{\name}[1][]{\textsc{CompRx}}
\usepackage{colortbl}
\newcommand{\stoptocwriting}{\addtocontents{toc}{\protect\setcounter{tocdepth}{-5}}}
\newcommand{\resumetocwriting}{\addtocontents{toc}{\protect\setcounter{tocdepth}{\arabic{tocdepth}}}}
\usepackage[toc,page,header]{appendix}
\usepackage{etoc}
\usepackage{tcolorbox}
\usepackage{enumitem}
\usepackage{booktabs}
\usepackage{tabularx}
\usepackage[table]{xcolor}
\usepackage{bibunits}
\definecolor{lightgray}{gray}{0.95}

\def\paperID{18581} % *** Enter the Paper ID here
\def\confName{CVPR}
\def\confYear{2026}

\title{CheXmix: Unified Generative Pretraining for Vision Language Models in Medical Imaging}

\author{
    Ashwin Kumar$^{1,2,3,\dagger}$ \enskip
    Robbie Holland$^{1,3}$ \enskip
    Corey Barrett$^2$ \enskip
    Jangwon Kim$^2$ \enskip
    Maya Varma$^{1,3}$ \enskip
    Zhihong \\ 
    Chen$^{1,3}$ \enskip
    Yunhe Gao$^{1,3}$ \enskip
    Greg Zaharchuk$^3$ \enskip
    Tara Taghavi$^2$ \enskip
    Krishnaram Kenthapadi$^2$ \enskip
    Akshay Chaudhari$^{1,3}$ \\[1ex]
    \normalsize $^1$Stanford AIMI, Stanford University \quad
    $^2$Oracle Health AI \quad
    $^3$Department of Radiology, Stanford University \\
}

\begin{document}
\maketitle

% This line forces the footnote to use your specific symbol
\renewcommand{\thefootnote}{\fnsymbol{footnote}}
\setcounter{footnote}{2} % 2 is the dagger symbol
\footnotetext{Corresponding author: akkumar@stanford.edu.}
\renewcommand{\thefootnote}{\arabic{footnote}} % Reset to numbers for other footnotes
\begin{abstract}

Recent medical multimodal foundation models are built as multimodal LLMs (MLLMs) by connecting a CLIP-pretrained vision encoder to an LLM using LLaVA-style finetuning. This two-stage, decoupled approach introduces a projection layer that can distort visual features. This is especially concerning in medical imaging where subtle cues are essential for accurate diagnoses. In contrast, early-fusion generative approaches such as Chameleon eliminate the projection bottleneck by processing image and text tokens within a single unified sequence, enabling joint representation learning that leverages the inductive priors of language models. We present CheXmix, a unified early-fusion generative model trained on a large corpus of chest X-rays paired with radiology reports. We expand on Chameleon's autoregressive framework by introducing a two-stage multimodal generative pretraining strategy that combines the representational strengths of masked autoencoders with MLLMs. The resulting models are highly flexible, supporting both discriminative and generative tasks at both coarse and fine-grained scales. Our approach outperforms well-established generative models across all masking ratios by 6.0\% and surpasses CheXagent by 8.6\% on AUROC at high image masking ratios on the CheXpert classification task. We further inpaint images over 51.0\% better than text-only generative models and outperform CheXagent by 45\% on the GREEN metric for radiology report generation. These results demonstrate that CheXmix captures fine-grained information across a broad spectrum of chest X-ray tasks. Our code is at: \href{https://github.com/StanfordMIMI/CheXmix}{https://github.com/StanfordMIMI/CheXmix}.

% can be less direct and potentially inefficient, as it does not optimize the entire pipeline end-to-end. 

% While LLaVA’s projection-based integration successfully bridges visual and linguistic modalities, it can , which may degrade classification performance and induce catastrophic forgetting. In medical imaging, this limitation is especially critical, since even a single misinterpreted visual cue can lead to incorrect diagnoses. In contrast, early-fusion generative pretraining approaches like Chameleon address these challenges by removing the projection bottleneck, instead processing 

% Several multimodal foundation models (FMs) for medical imaging have relied on CLIP-style pretraining. We here explore \textit{generative pretraining} by tokenizing images and treating multimodal data as a unified sequence of tokens, leveraging the strong inductive priors of language models. We develop unified early-fusion generative models trained on a large-corpus of chest X-rays paired with radiology reports. The resulting models are flexible, supporting both discriminative and generative tasks. Our approach outperforms leading contrastive methods by 14.3\% on the CheXpert classification task at high image masking ratios, inpaints images 51.1\% better than text-only generative models, and generates radiology reports more accurately than a general-domain multimodal generative model. These results demonstrate that our unified multimodal generative pretraining framework captures fine-grained information across both discriminative and generative medical imaging tasks.
\end{abstract}
\stoptocwriting
\begin{figure}[ht]
\centerline{\includegraphics[width=0.5\textwidth]{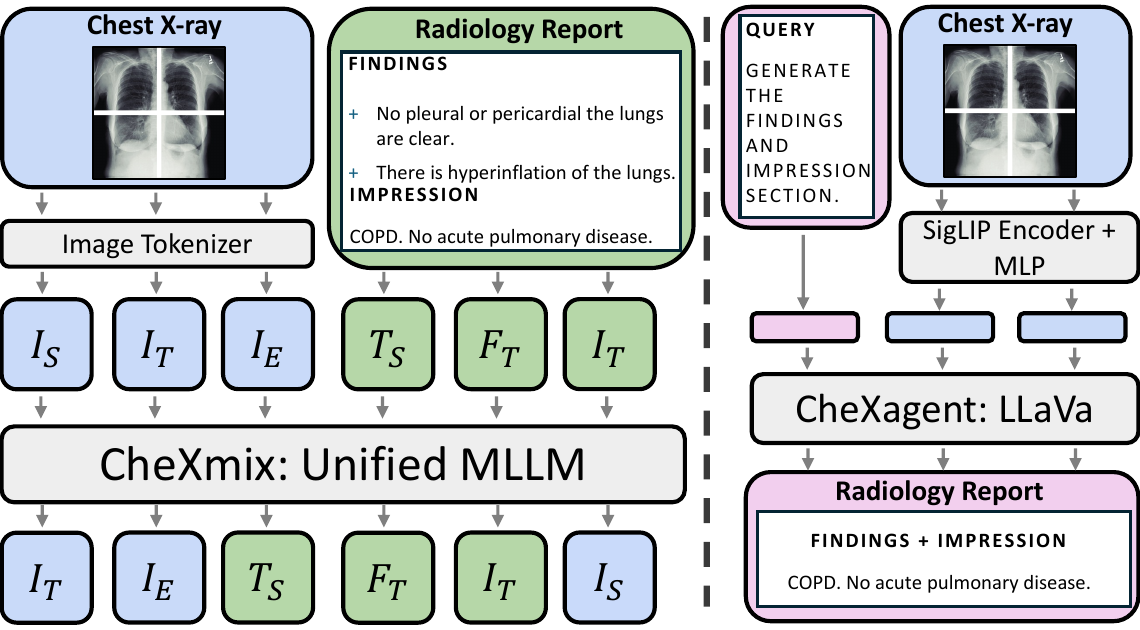}}
\caption{\textbf{Architectural comparison of CheXmix and CheXagent.} Architectural and functional differences between our proposed model, CheXmix, and the LLaVA-style model, CheXagent. CheXmix, a unified early-fusion generative model, natively offers report generation capabilities directly after pretraining. In contrast, CheXagent requires full instruction finetuning for generative tasks and utilizes a separate SigLIP encoder for discriminative functions. For classification, CheXmix directly uses its learned image embeddings, while CheXagent relies on a distinct pretrained SigLIP encoder. CheXmix's modular pretraining strategy yields strong performance across both discriminative and generative medical imaging tasks, demonstrating better flexibility. Abbreviations: 
$I_S / I_E =$ start / end image token, 
$I_T =$ image token, 
$T_S =$ text start token, 
$F_T / I_T =$ findings / impression token.}
\label{fig:cxagent_vs_cxmix}
\end{figure}

\section{Introduction}
\label{sec:intro}

\begin{figure*}[ht]
% \thisfloatpagestyle{empty}
\centerline{\includegraphics[width=1.0\textwidth]{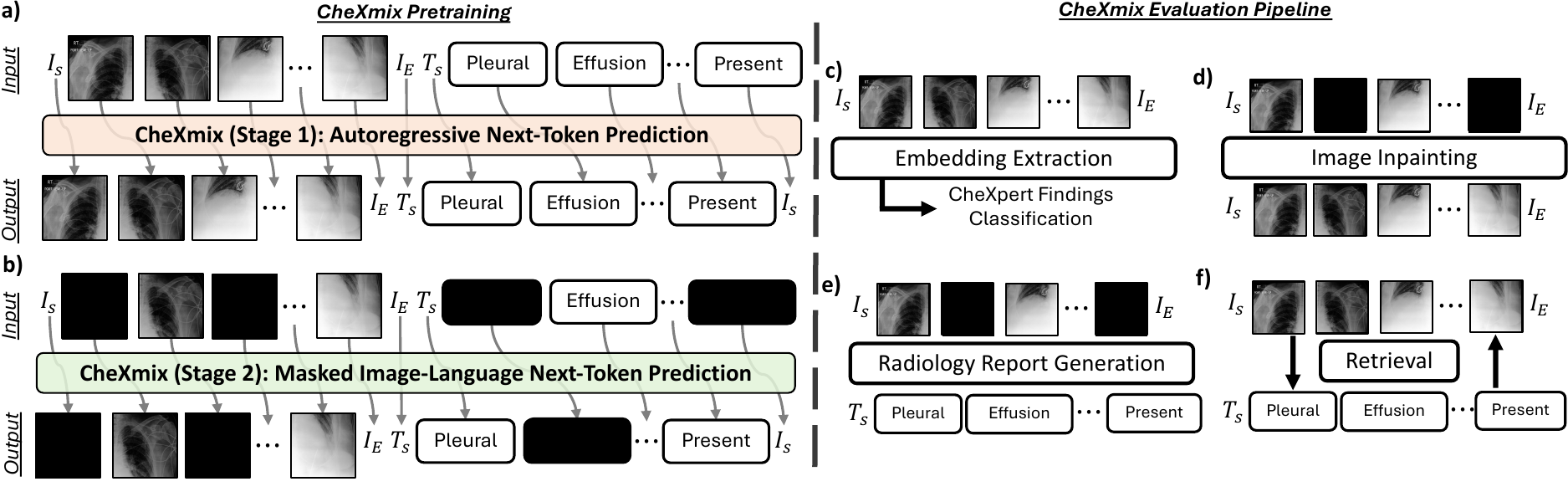}}
\caption{\textbf{CheXmix Generative Pretraining Overview} (a) Chest X-rays are tokenized using VQ-GAN, and text is tokenized with the RadPhi-2 tokenizer. The RadPhi-2 transformer decoder model is trained for next-token prediction, with special tokens $I_S$ (Image Start), $I_E$ (Image End), and $T_S$ (Text Start). (b) During training, 50\% of image and text tokens are masked, and the next-token prediction loss is computed only for unmasked outputs. (c)-(e) After pretraining, CheXmix flexibly generates text or image tokens, and we evaluate it on (c) CheXpert embedding findings classification, (d) image inpainting, (e) radiology report generation, and (f) retrieval.}
\label{fig:overview}
\end{figure*}

Medical imaging, such as X-rays, CT scans, and MRIs, plays a central role in diagnosing and monitoring patient health. This imaging data is inherently multimodal, often accompanied by corresponding radiology reports, making it well suited for vision-language multimodal training. Recent multimodal medical imaging foundation models (FMs)~\citep{chexagent-2024, hamamci2024developing, blankemeier_kumar2026merlin} have predominantly employed contrastive learning strategies~\citep{radford2021learning}. The resulting vision encoders are then integrated into multimodal architectures following the LLaVA paradigm~\citep{liu2023visual, liu2024improved} to form multimodal large language models (MLLMs), where pretrained visual features are projected into the LLM’s input space through a linear or MLP adapter. Despite its effectiveness, this paradigm faces significant limitations. Visual signals may be ineffectively translated for the LLM decoder through the projection layer~\citep{lin2024video, verma2024cross}, potentially impacting an MLLM's performance on classification tasks compared to specialized discriminative models~\cite{zhang2024visually}. CLIP features are also optimized for contrastive objectives and do not necessarily transfer well to all downstream tasks, especially generative applications~\cite{kang2025clip, li2024erroneous}. Furthermore, multimodal instruction tuning can induce catastrophic forgetting, where the model's linguistic capabilities degrade below those of its base LLM~\citep{srivastava2024improving, he2023continual}, which can hinder the model's flexibility and generalizability. Accurate visual feature input is critical in the medical domain, where a single, fine-grained visual cue can indicate a specific condition.

These challenges motivate the exploration of alternative unified pretraining strategies that offer more seamless multimodal integration. General-domain early-fusion multimodal generative models, e.g. Chameleon~\citep{team2024chameleon}, tokenize images at a patch level alongside text and process image-text data as a unified sequence of tokens. Since these images are compressed through a VQ-GAN tokenizer~\citep{team2024chameleon}, unlike LLaVa's continuous CLIP embeddings, these discrete tokens can be well adapted natively into an LLM's vocabulary while still retaining patch-level visual information. By avoiding a pretrained image encoder and training on a large data corpus, Chameleon's transformer decoder model learns joint, multimodal image-text representations from scratch.

Building on this paradigm, we present CheXmix, a unified early-fusion generative transformer decoder model that mixes image and text tokens within a shared token sequence (Figure \ref{fig:cxagent_vs_cxmix}). We pretrain CheXmix on a large corpus of chest X-rays paired with radiology reports. To enable this, we curate a dataset comprising over 627 million tokens across five public chest X-ray datasets. Chest X-rays offer the largest available collections of paired image–text data in medical imaging, making them an ideal testbed for developing early-fusion pretraining approaches. Our work focuses on developing and evaluating CheXmix’s pretraining, highlighting the inherent adaptability and representational flexibility of early-fusion generative pretraining in the medical domain.

Specifically, CheXmix uses a two-stage multimodal generative pretraining approach combining the strengths of MAEs and MLLMs through standard and masked autoregressive pretraining. MAE strategies improve fine-grained visual representations on both discriminative and generative tasks by reconstructing missing image patches \citep{he2022masked}. We apply masking to both image and text tokens, creating a strong generative objective that facilitates joint representation learning in autoregressive models. 

We evaluate CheXmix across both discriminative and generative tasks to assess whether our generative pretraining strategy yields suitable representations for chest X-ray tasks broadly. Prior early-fusion generative models~\citep{team2024chameleon, xie2024show, zhou2024transfusion} have not systematically isolated or analyzed their pretrained image representations at the embedding level. Understanding how well these unified models encode medical visual information provides insight into the visual features learned through joint multimodal pretraining. Furthermore, this pretraining paradigm offers inherent flexibility in the medical imaging domain, enabling diverse downstream capabilities such as report generation and image inpainting. To assess this flexibility, we evaluate CheXmix’s ability to generate and inpaint images, tasks that probe the model’s capacity to capture both semantic and spatial structure. We demonstrate these advantages by evaluating our CheXmix models (Figure~\ref{fig:overview}a–b) on CheXpert embedding-level findings classification, image inpainting, radiology report generation, and retrieval (Figure~\ref{fig:overview}c–f).

% Successful inpainting of anatomical regions demonstrates that the model learns meaningful, generalizable representations. 

In this paper, we provide the following contributions: 
\begin{enumerate}
    \item We introduce \textit{CheXmix}, a multimodal generative pretraining strategy that mixes image and text tokens in an interleaved sequence to jointly represent chest X-rays and radiology reports. Our approach employs a masked image-language pretraining strategy that integrates techniques from MAEs and MLLMs, enhancing the robustness of CheXmix's learned representations. Compared to standard next-token prediction, this masking strategy yields substantial gains: 6.7\% improvement in CheXpert classification, 20\% in inpainting images, and 56\% in radiology report generation at higher masking ratios (Tables ~\ref{tab:chexpert_auroc_std}, ~\ref{tab:image_quality_metrics}, ~\ref{tab:rrg_main}). 
    \item CheXmix (S1 + S2) outperforms established generative models by 6.0\% in AUROC and exceeds CheXagent by 8.6\% at higher masking ratios (Table ~\ref{tab:chexpert_auroc_std}). For radiology report generation, CheXmix (S1 + S2) surpasses CheXagent by 45.0\% on the GREEN metric while achieving comparable CheXbert scores (Table ~\ref{tab:rrg_main}). These results highlight that CheXmix's unified architecture could serve as a viable alternative to LLaVA-style finetuning in the medical domain.
    \item We demonstrate CheXmix’s architectural flexibility by using test-time augmentation (TTA) to improve report generation by nearly 13\% on average, without any additional pretraining (Figure ~\ref{fig:test_time_ensembling}).
\end{enumerate}
\section{Related work}
\label{sec:related_work}

\paragraph{General Domain Multimodal Pretraining}
Seminal work on multimodal pretraining has been largely dominated by contrastive learning, with models such as CLIP~\cite{radford2021learning} and its variants like SigLIP~\cite{zhai2023sigmoid} demonstrating strong discriminative performance across tasks including retrieval, linear probing, and zero-shot classification. To extend these pretrained representations for generative capabilities, approaches such as LLaVA~\cite{liu2023visual} and BLIP-2~\cite{li2023blip} freeze these encoders and connect them to a LLM using a lightweight projection layer or adapter. However, visual signals may be imperfectly translated through the projection layer~\citep{verma2024cross}, and multimodal instruction tuning can induce catastrophic forgetting~\citep{srivastava2024improving, he2023continual}, both of which can limit an MLLM’s performance on discriminative tasks~\cite{zhang2024visually} and reduce its flexibility and generalizability.

\paragraph{Medical Multimodal Pretraining}
In the medical domain, multimodal pretraining has largely mirrored these general-domain strategies. On one hand, multimodal contrastive models such as GLoRIA~\cite{huang2021gloria}, BiomedCLIP~\cite{zhang2023biomedclip}, and Merlin~\cite{blankemeier_kumar2026merlin} learn joint embeddings of medical images and corresponding text, such as radiology reports, and perform well on discriminative tasks including zero-shot classification and retrieval. On the other hand, LLaVa-style connector methods, including LLaVA-Med~\cite{li2023llava} and Med-Palm~\cite{tu2024towards}, adapt general-domain multimodal LLMs for medical visual question answering and report generation, but require extensive instruction fine-tuning. State-of-the-art models for chest X-ray understanding and report generation, such as CheXagent~\cite{chexagent-2024}, first finetune the SigLIP image and text encoders and then develop a LLaVa-style LLM for generative tasks using large-scale instruction datasets. However, these approaches typically result in specialized models, one for discriminative tasks using embeddings and another for generative tasks using the LLM decoder, leaving a gap for a single, flexible model that can effectively handle both types of tasks.

% \todo{Prob not space but see at end: There could even be a small section on text to image models and how an approach like this could be a bridge to such diffusion models?}

\paragraph{Unified Generative Multimodal Pretraining.}
General-domain early-fusion multimodal generative models, such as Chameleon~\citep{team2024chameleon}, tokenize images alongside text and process image–text data as a unified sequence of tokens. Building on this approach, recent work in the medical domain has begun to adapt unified multimodal pretraining to clinical tasks; for example, ProgEMU~\cite{ma2025towards} translates the EMU~\cite{sun2023emu} architecture to chest x-ray applications. Specifically, ProgEMU focuses on training a transformer decoder for progression-aligned counterfactual generation rather than joint image-report modeling. In the general domain, Transfusion~\cite{zhou2024transfusion} and Show-O~\cite{xie2024show} surpass Chameleon by combining diffusion objectives for images with autoregressive modeling for text. CheXmix follows the early-fusion principle but applies it to radiology by jointly modeling a chest X-ray and its report within a single generative sequence, eliminating the need for CLIP-style contrastive alignment.

% , whereas our goal is to jointly learn a chest X-ray with its corresponding radiology report in a single sequence

% These models differ primarily in their image representation: Transfusion operates on continuous vectors, whereas Show-O uses discrete tokens with discrete denoising diffusion. Notably, both employ full bidirectional attention over image sequences to improve visual representation.

\paragraph{Masked Modeling in Unified Architectures.}  
MAEs~\cite{he2022masked} have demonstrated that random masking provides a powerful self-supervised signal for learning robust visual representations, particularly when leveraged by Vision Transformers (ViTs) with bidirectional attention. In the medical domain, M3AE~\cite{chen2022multi} demonstrated the effectiveness of masking both chest x-rays and radiology reports, achieving state-of-the-art performance across multiple benchmarks. However, M3AE relies on a dual-encoder architecture with a cross-modal fusion module. Consequently, it is not a natively unified generative model; adapting it for generative tasks requires grafting separate task-specific decoders or employing LLaVA-style projection layers to connect it to an LLM. Recently, general-domain unified models like Show-O and Transfusion have revisited full, bidirectional attention for image modeling within decoder-only transformers. Building on this, we train CheXmix both with and without causal image masking to assess how bidirectional and causal attention affect unified multimodal learning in the medical domain (Table ~\ref{tab:causal_mask_modeling}).
\section{Method}
\label{sec:method}

\subsection{Dataset Pre-processing}
\label{method:dataset_processing}

To construct a training dataset for generative pretraining, we aggregate five publicly available chest X-ray datasets from diverse institutions: MIMIC-CXR \citep{johnson2019mimic}, CheXpert \citep{irvin2019chexpert, chambon2024chexpert}, PadChest \citep{bustos2020padchest}, BIMCV-COVID19 \citep{vaya2020bimcv}, and OpenI \citep{shih2019augmenting}. Each image is paired with its corresponding radiology report, including both the findings and impression sections. Images are encoded into 1024 discrete tokens using Chameleon's VQ-GAN tokenizer \citep{team2024chameleon} and text is tokenized with the RadPhi-2 text tokenizer ($|\mathcal{V}| =$ 50,368) \citep{chexagent-2024}. This preprocessing yields 550,395 image-text training pairs and 14,111 test pairs, resulting in a total of 627,809,814 tokens, comprising 577,054,144 image tokens and 49,755,670 text tokens. Over 99\% of our image-text sequences contained 1,298 or fewer tokens in the training set and 1,295 in the test set. We further cap the context length at 1,300 tokens, keeping all image tokens intact while cropping text tokens, with over 99\% of sequences fitting this limit. 

% decoder-only transformer with 32 layers, 32 attention heads per layer, and a maximum context length of 2,048 tokens. It was 

\subsection{Model Pretraining Approach}
\label{method:model_pretraining}

We initialize CheXmix using the RadPhi-2 language decoder \citep{chexagent-2024}, a language model with comprehensive medical and clinical knowledge. RadPhi-2 is adapted from Phi-2 \citep{li2023textbooks}, a 2.7B-parameter model trained on a clinical text corpus (over 2.7T tokens) using next-token prediction as the training objective.

In this work, we introduce a two-stage multimodal training approach that leverages the medical inductive prior of RadPhi-2, starting with standard autoregressive pretraining on image and text tokens, followed by masked autoregressive pretraining to encourage fine-grained representation learning. Additional training configurations are provided in Supplementary Section \ref{appendix:section_a} and \ref{appendix:section_b1}.

\paragraph{Stage 1: Standard Autoregressive Pretraining:} 

From our large multimodal image-text pretraining corpus $\mathcal{D} = \{(x_i^{\text{img}}, x_i^{\text{text}})\}_{i=1}^K$, we obtain tokenized image representations, effectively treating each image as a series of discrete patch tokens to be processed by the language decoder alongside text tokens. Each chest X-ray ($x^{\text{img}}$) is tokenized into a sequence of 1,024 discrete image tokens $\mathbf{z} = (z_1, \dots, z_{1024})$ from a codebook of size 8,192 using a VQ-GAN tokenizer \citep{team2024chameleon}. Prior work has demonstrated that off-the-shelf VQ-GAN models maintain strong encoding and reconstruction performance on medical imaging tasks without retraining~\citep{chambon2022adapting, varma2025medvae}. The corresponding text sequence ($x^{\text{text}}$), consisting of the findings and impressions sections from the radiology report, is tokenized with the RadPhi-2 text tokenizer into $\mathbf{y} = (y_1, \dots, y_m)$. Image and text tokens are then combined into a joint sequence of length $N$, either \resizebox{\columnwidth}{!}{$
\mathbf{S} = (z_1, \dots, z_{1024}, y_1, \dots, y_m) \quad \text{or} \quad \mathbf{S} = (y_1, \dots, y_m, z_1, \dots, z_{1024})
$} and the order randomized such that the image precedes the text in 50\% of cases \citep{team2024chameleon}. To mark modality boundaries, we prepend special tokens: image sequences are wrapped with start and end markers, while text sequences begin with a start token (Figure ~\ref{fig:overview}). The joint vocabulary is expanded accordingly, yielding $|\mathcal{V}| = 58{,}592 = |\mathcal{V}_{\text{text}}| + |\mathcal{V}_{\text{img}}|$. Following the RadPhi-2 setup, we apply autoregressive next-token prediction (Equation ~\ref{eq:unified_ntp}), now applied to both image and text tokens (Figure ~\ref{fig:overview}a). We train for over 700,000 steps with an effective batch size of 32 on four NVIDIA A100 GPUs (80 GB).

\begin{equation} \label{eq:unified_ntp}
\mathcal{L}_{\text{NTP}} = - \sum_{i=1}^{N} \log p_{\theta}(s_i | s_1, \dots, s_{i-1})
\end{equation}

\paragraph{Stage 2: Masked Image-Language Pretraining:}

After autoregressive pretraining, we perform a second training step using random masking to further improve the discriminative and generative capabilities of the model. During this stage, we create a corrupted input sequence $\textbf{$\mathbf{S}'$} = (s_1, s', \dots, s', s)$ by randomly replacing 50\% of image and text tokens in the original sequence \textbf{$S$} with a special \texttt{[MASK]} token, denoted as $s'$. As a result, the model receives a mixed input sequence of unmasked tokens and masked tokens from the set $\mathcal{M}$, e.g., $\mathbf{S}' = (s_1, \texttt{[MASK]}, s_3, \texttt{[MASK]}, \dots, s_N)$.

While the model processes this full sequence to build context, we apply the autoregressive masked loss (Equation~\ref{eq:mil}) only to the masked positions. This forces the model to reconstruct the missing information based on previous unmasked and masked tokens (Figure ~\ref{fig:overview}b), effectively learning to generate missing information. We train for over 500,000 steps with an effective batch size of 32 on four NVIDIA A100 GPUs (80 GB).

\begin{equation} \label{eq:mil}
\mathcal{L}_{\text{MIL}} = - \sum_{i \in \mathcal{M}} \log p_{\theta}(s_i \mid \mathbf{S}'_{<i})
\end{equation}

\subsection{Evaluation}

We evaluate CheXmix's pretrained representations on both discriminative and generative tasks. Discriminative performance is measured via CheXpert findings classification, comparing embeddings to relevant general-domain and medical-specific baselines. Generative capabilities are assessed through image inpainting and radiology report generation across \textit{five} multimodal chest X-ray datasets (Section~\ref{method:dataset_processing}). To probe robustness, we examine multiple masking ratios (20\%–80\%) and evaluate test-time augmentation strategies for report generation that require no additional training. Additional methods regarding our evaluation strategy are provided in Supplementary Section \ref{appendix:section_b2}--\ref{appendix:section_b4}.

\begin{table*}[htbp]
\centering
\small
\resizebox{\textwidth}{!}{%
\begin{tabular}{c|ccccccc}
\toprule
\multicolumn{7}{c|}{\textbf{Generative Pretraining Objective}} & \multicolumn{1}{c}{\textbf{Reference Maximum}} \\
\midrule
\textbf{Masking \%} & \textbf{Chameleon} & \textbf{HealthGPT} & \textbf{MAE} & \textbf{M3AE} & \textbf{CheXmix (S1)} & \textbf{CheXmix (S1 + S2)} & \textbf{CheXagent} \\
\midrule
0\% & 0.695 \tiny{$\pm$0.000} & 0.641 \tiny{$\pm$0.002} & 0.641 \tiny{$\pm$0.001} & 0.676 \tiny{$\pm$0.000} & 0.664 {\tiny ±0.000} & \underline{0.712 \tiny{$\pm$0.001}} & \textbf{0.831 \tiny{$\pm$0.001}} \\
20\% & 0.674 \tiny{$\pm$0.001} & 0.639 \tiny{$\pm$0.002} & 0.634 \tiny{$\pm$0.003} & 0.663 \tiny{$\pm$0.000} & 0.660 \tiny{$\pm$0.000} & \underline{0.705 \tiny{$\pm$0.001}} & \textbf{0.756 \tiny{$\pm$0.002}} \\
40\% & 0.649 \tiny{$\pm$0.001} & 0.566 \tiny{$\pm$0.007} & 0.626 \tiny{$\pm$0.000} & 0.655 \tiny{$\pm$0.000} & 0.653 {\tiny ±0.001} & \textbf{0.702 \tiny{$\pm$0.001}} & 0.683 \tiny{$\pm$0.005} \\
60\% & 0.618 \tiny{$\pm$0.001} & 0.563 \tiny{$\pm$0.005} & 0.637 \tiny{$\pm$0.003} & 0.648 \tiny{$\pm$0.001} & 0.643 {\tiny ±0.000} & \textbf{0.689 \tiny{$\pm$0.002}} & 0.634 \tiny{$\pm$0.008} \\
80\% & 0.592 \tiny{$\pm$0.000} & 0.562 \tiny{$\pm$0.003} & 0.621 \tiny{$\pm$0.002} & 0.627 \tiny{$\pm$0.000} & 0.599 {\tiny ±0.000} & \textbf{0.656 \tiny{$\pm$0.002}} & 0.569 \tiny{$\pm$0.003} \\
\bottomrule
\end{tabular}%
}
\caption{\textbf{CheXpert Embedding Findings Classification.} CheXmix (S1 + S2) consistently outperforms other generative models at all masking percentages, demonstrating improved robustness under occlusion. \textbf{Bold} indicates the best-performing model for each masking percentage, while \underline{underline} marks the best-performing generative model. AUROC (mean~$\pm$~std) is reported across three random seeds, with standard deviation computed as the average over seeds.}
\label{tab:chexpert_auroc_std}
\end{table*}

\section{Results}
We rigorously evaluate CheXmix’s representational quality through both discriminative and generative tasks (Figure \ref{fig:overview}c–f), outperforming general-domain and medical-specific baselines, particularly at high masking ratios. Evaluations were conducted across 20\%, 40\%, 60\%, and 80\% masking to probe CheXmix’s fine-grained representational capabilities. We further demonstrate CheXmix’s flexibility using test-time augmentation (TTA) to improve report generation without additional training. Ablations are provided to motivate our hyperparameter choices. Extended results and ablations are reported in Supplementary Section \ref{appendix:extended_results}.

\label{sec:results}

\subsection{CheXpert Classification Task}

We evaluate CheXmix's pretrained image representations on the 14 CheXpert findings (Table \ref{tab:chexpert_auroc_std}) across multiple masking levels to assess robustness under partial visual information. This setup also enables us to isolate the contribution of CheXmix's pretraining stages by comparing S1 versus S1 + S2. As generative baselines, we include: \textit{Chameleon} \citep{team2024chameleon}, a 7B early-fusion generative model trained on 4.8T general-domain image-text tokens; HealthGPT~\cite{lin2025healthgptmedicallargevisionlanguage}, an early-fusion medical-specific generalist baseline; \textit{MAE~\citep{he2022masked}}, a general-domain masked autoencoder trained on ImageNet; and \textit{M3AE~\citep{chen2022multi}}, a multimodal masked autoencoder trained on chest X-rays and radiology reports. We additionally include \textit{CheXagent's SigLIP} image encoder~\citep{chexagent-2024}, a strong vision encoder pretrained on large-scale natural and medical imaging datasets, serving as a high-performance reference point. For additional details, including the use of hidden layers from these models for this task, please see Supplementary Section \ref{text:chexpert_findings_cls}.

At 0\% masking, CheXmix (S1 + S2) outperforms CheXmix (S1), M3AE, MAE, and Chameleon by 7.2\%, 5.3\%, 11.0\%, and 2.4\%, respectively, establishing it as the strongest generatively pretrained model for discriminative representations. 

% However, its performance remains below that of CheXagent. 

We additionally evaluate all baselines under 20\%, 40\%, 60\%, and 80\% image masking. Masking during evaluation acts as a controlled probe of representation quality ~\cite{pathak2016context}, requiring embeddings learned under heavy masking to capture semantic features. Across masking levels, CheXmix (S1 + S2) surpasses CheXmix (S1) by 6.7\% (average AUROC), underscoring the contribution of masked image-language generative pretraining beyond standard next-token prediction for learning robust, fine-grained features. CheXmix (S1 + S2) further outperforms M3AE, MAE, and Chameleon by 6.2\%, 9.4\%, and 8.7\% across all masking ratios. Although M3AE also uses a multimodal generative masking objective, our unified early-fusion generative approach with masked image-language pretraining consistently yields stronger representations. Notably, CheXmix surpasses Chameleon despite Chameleon being trained on a dataset roughly 8,000× larger with a model nearly 3× the size, underscoring the value of domain-specific unified generative pretraining. Additionally, while SigLIP performs better than CheXmix (S1 + S2) pretraining at low masking ratios (0–20\%), CheXmix (S1 + S2) outperforms SigLIP by 8.6\% at higher masking ratios (40-80\%), showing better fine-grained feature retention under occlusion. 

Collectively, these results show CheXmix's multi-stage domain-specific masked multimodal generative pretraining substantially improves fine-grained discriminative performance. This allows CheXmix to outperform larger general-domain models, medical multimodal generatively pretrained models, and medical discriminative encoders.

% These results collectively demonstrate the benefits of domain-specific finetuning for generative models at the embedding level, the advantages of multimodal pretraining for fine-grained tasks requiring image embeddings, and the ability of masked autoregressive pretraining to outperform much larger natural image models when fine-grained detail is critical.

\subsection{Image Inpainting}

To evaluate fine-grained generative performance, we design an inpainting task on chest X-rays, where models must reconstruct masked image regions. Successful inpainting probes representation quality~\citep{pathak2016context} and indicates that the model learns meaningful, generalizable features.

\begin{figure}[ht]
\centerline{\includegraphics[width=0.5\textwidth]{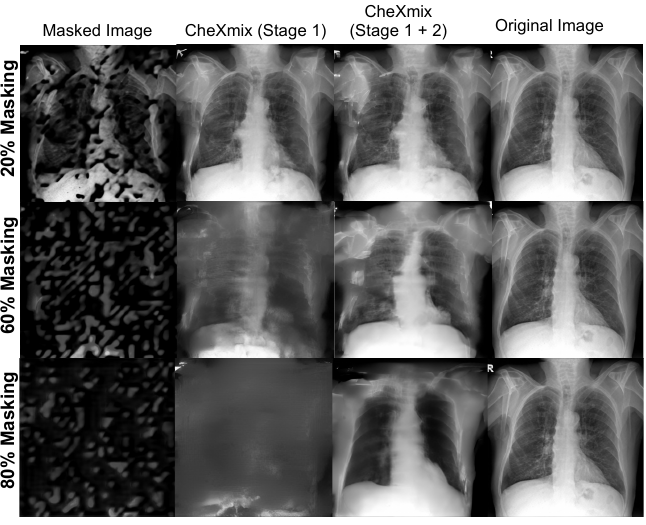}}
\caption{\textbf{Image Inpainting Visualization} CheXmix (S1 + S2) pretraining shows considerable image inpainting improvement at higher masking ratios.}
\label{fig:image_quality_qualitative}
\end{figure}

As baselines we include VQ-GAN, in which masked tokens are left blank and decoded without modification, and RadPhi-2. As shown in (Table \ref{tab:image_quality_metrics}), RadPhi-2 outperforms VQ-GAN, achieving a 80.0\% improvement in PSNR across masked regions. CheXmix (S1) further surpasses RadPhi-2 by 26.0\% in PSNR across masking percentages. By incorporating masking during pretraining, CheXmix (S1 + S2) offers an additional 20.0\% improvement over stage 1 and a 51.1\% gain over RadPhi-2 in PSNR across masked regions. These findings highlight that incorporating multimodal image information during pretraining substantially enhances image generation quality, with the benefits of masking evident in both quantitative metrics (Table ~\ref{tab:image_quality_metrics}) and qualitative reconstructions (Figure \ref{fig:image_quality_qualitative}).

Therefore, our strong performance on this task using our masked autoregressive model demonstrates that our pretraining approach learns useful and generalizable representations, supporting the central claim of our work.

\begin{table*}[htbp]
\centering
\small
\resizebox{\textwidth}{!}{% Changed resize factor back
%
% 1. Removed 'cc|' from the column definition
%
\begin{tabular}{c|cc|cc|cc|cc}
\toprule
%
% 2. Removed the MAE column header
%
\multirow{2}{*}{\textbf{Masking \%}} & \multicolumn{2}{c|}{\textbf{VQ-GAN}} & \multicolumn{2}{c|}{\textbf{RadPhi-2}} & \multicolumn{2}{c|}{\textbf{CheXmix (S1)}} & \multicolumn{2}{c}{\textbf{CheXmix (S1 + S2)}} \\
%
% 3. Adjusted cmidrule ranges
%
\cmidrule(lr){2-3} \cmidrule(lr){4-5} \cmidrule(lr){6-7} \cmidrule(lr){8-9}
%
% 4. Removed the MAE metrics header
%
 & \textbf{PSNR}$\uparrow$ & \textbf{MS-SSIM}$\uparrow$ & \textbf{PSNR}$\uparrow$ & \textbf{MS-SSIM}$\uparrow$ & \textbf{PSNR}$\uparrow$ & \textbf{MS-SSIM}$\uparrow$ & \textbf{PSNR}$\uparrow$ & \textbf{MS-SSIM}$\uparrow$ \\
\midrule
\addlinespace
%
% 5. Removed MAE data from all rows
%
20 & 10.61{\scriptsize$\pm$0.01} & 0.482{\scriptsize$\pm$0.00} & 18.98{\scriptsize$\pm$0.04} & 0.735{\scriptsize$\pm$0.00} & \textbf{24.13{\scriptsize$\pm$0.03}} & \textbf{0.867{\scriptsize$\pm$0.00}} & \textbf{24.13{\scriptsize$\pm$0.03}} & 0.859{\scriptsize$\pm$0.00} \\
% \addlinespace
40 & 7.18{\scriptsize$\pm$0.01} & 0.329{\scriptsize$\pm$0.00} & 13.47{\scriptsize$\pm$0.05} & 0.582{\scriptsize$\pm$0.00} & 19.78{\scriptsize$\pm$0.03} & 0.757{\scriptsize$\pm$0.00} & \textbf{21.25{\scriptsize$\pm$0.03}} & \textbf{0.768{\scriptsize$\pm$0.00}} \\
% \addlinespace
60 & 6.29{\scriptsize$\pm$0.02} & 0.252{\scriptsize$\pm$0.00} & 11.08{\scriptsize$\pm$0.05} & 0.485{\scriptsize$\pm$0.00} & 13.84{\scriptsize$\pm$0.02} & 0.607{\scriptsize$\pm$0.00} & \textbf{19.25{\scriptsize$\pm$0.03}} & \textbf{0.699{\scriptsize$\pm$0.00}} \\
% \addlinespace
80 & 5.91{\scriptsize$\pm$0.02} & 0.249{\scriptsize$\pm$0.00} & 10.38{\scriptsize$\pm$0.04} & 0.418{\scriptsize$\pm$0.00} & 10.24{\scriptsize$\pm$0.02} & 0.455{\scriptsize$\pm$0.00} & \textbf{16.96{\scriptsize$\pm$0.03}} & \textbf{0.632{\scriptsize$\pm$0.00}} \\ 
\bottomrule
\end{tabular}}
\caption{\textbf{Image Inpainting Quantitative Results:} Multimodal generative pretraining improves inpainting performance, with CheXmix (S1 + S2) showing notable advantages at higher masking percentages, demonstrating greater robustness to fine-grained perturbations in generative tasks (best metrics in \textbf{bold}). We compute PSNR and MS-SSIM on a random sample of 5,000 images and report mean and standard deviation across three runs with different random seeds. For RadPhi-2 and CheXmix models, inpainting is performed on image tokens generated by the VQ-GAN image tokenizer.}
\label{tab:image_quality_metrics} % Keeping the label from previous edit
\end{table*}

\subsection{Radiology Report Generation}

% Automatic radiology report generation is a particularly challenging task for deep learning models, as reports must capture fine-grained clinical details while remaining logically consistent and free of hallucinated findings~\citep{wang2024survey, reiner2009challenges}. Unlike general NLP tasks, linguistic similarity alone is insufficient, generated reports must also be clinically accurate, which traditional metrics like BLEU or ROUGE fail to capture~\cite{ostmeier2024green}.

Automatic radiology report generation is challenging for deep learning models, requiring fine-grained clinical accuracy while avoiding hallucinated or inconsistent findings~\citep{wang2024survey, reiner2009challenges}. We evaluate radiology report generation (Table~\ref{tab:rrg_main}) by providing models with varying levels of masked image inputs, allowing us to assess their ability to produce reports under incomplete visual information. For evaluation, we use GREEN, a metric designed to measure clinical factuality and hallucination detection, alongside CheXbert, which assesses the correctness of structured disease labels in the generated reports~\cite{ostmeier2024green, smit2020CheXbert}. Additional metrics and model evaluations can be found in Supplementary Section \ref{appendx:rrg_quantative_extended} and qualitative examples of model-generated radiology reports can be found in Supplementary Section \ref{appendix:section_d2}.

We compare CheXmix against CheXagent, a state-of-the art foundation model for chest x-ray understanding and report generation, and Chameleon. Without any masking, CheXmix (S1) and CheXmix (S1 + S2) perform comparably on both GREEN and CheXbert. CheXmix outperforms CheXagent by 45.0\% on GREEN with relatively comparable performance on CheXbert. Compared to Chameleon, CheXmix (S1 + S2) achieves an order-of-magnitude improvement on GREEN and a 93.0\% improvement on CheXbert. Across masking ratios, CheXmix (S1 + S2) demonstrates robust performance, with only a 25.0\% decrease in GREEN from 0–80\% masking, whereas CheXmix (S1) experiences a 329.0\% drop.  At higher masking ratios, CheXmix (S1 + S2) continues to outperform CheXagent by over 45.0\% on GREEN and 42.0\% on CheXbert. These results highlight the benefits of CheXmix’s generative pretraining for producing accurate radiology reports under partial visual information.

\paragraph{Test-Time Augmentation:} We also investigate test-time augmentation (TTA) to improve report generation using CheXmix (S1 + S2) \textit{without} additional training (Figure ~\ref{fig:test_time_ensembling}). Specifically, we convert a single image token sequence into five disjoint masked sequences at 20\% masking and five disjoint unmasked sequences at 80\% masking. Using TTA, we observe a 10\% improvement on both GREEN and CheXbert at 20\% masking. We also notice a nearly 10\% and over 25\% improvement on GREEN and CheXbert, respectively, at 80\% masking. Using test-time augmentation with 20\% masked image tokens, report generation improves by 11.0\% on the GREEN metric compared to generating reports from unmasked images. These results demonstrate that CheXmix (S1 + S2) can leverage TTA to improve report quality post-training.

\begin{table*}[htbp]
\centering
\small
\resizebox{\textwidth}{!}{
\begin{tabular}{c|cc|cc|cc|cc}
\toprule
\multirow{2}{*}{\textbf{Masking \%}}
& \multicolumn{2}{c|}{\textbf{Chameleon}}
& \multicolumn{2}{c|}{\textbf{CheXagent}}
& \multicolumn{2}{c|}{\textbf{CheXmix (S1)}}
& \multicolumn{2}{c}{\textbf{CheXmix (S1 + S2)}} \\
\cmidrule(lr){2-3} \cmidrule(lr){4-5} \cmidrule(lr){6-7} \cmidrule(lr){8-9}
& \textbf{GREEN}$\uparrow$ & \textbf{CheXbert}$\uparrow$
& \textbf{GREEN}$\uparrow$ & \textbf{CheXbert}$\uparrow$
& \textbf{GREEN}$\uparrow$ & \textbf{CheXbert}$\uparrow$
& \textbf{GREEN}$\uparrow$ & \textbf{CheXbert}$\uparrow$ \\
\midrule
0
& 0.019 {\tiny ±0.005} & 0.202 {\tiny ±0.018}
& 0.152 {\tiny ±0.011} & 0.383 {\tiny ±0.022}
& 0.219 {\tiny ±0.015} & \textbf{0.390 {\tiny ±0.022}}
& \textbf{0.221 {\tiny ±0.015}} & \textbf{0.390 {\tiny ±0.023}} \\
% \addlinespace
20
& 0.039 {\tiny ±0.007} & 0.178 {\tiny ±0.018}
& 0.134 {\tiny ±0.009} & 0.357 {\tiny ±0.023}
& 0.181 {\tiny ±0.013} & \textbf{0.396 {\tiny ±0.022}}
& \textbf{0.215 {\tiny ±0.015}} & \textbf{0.396 {\tiny ±0.023}} \\
% \addlinespace
40
& 0.022 {\tiny ±0.004} & 0.171 {\tiny ±0.019}
& 0.117 {\tiny ±0.010} & 0.343 {\tiny ±0.022}
& 0.148 {\tiny ±0.012} & 0.376 {\tiny ±0.022}
& \textbf{0.224 {\tiny ±0.015}} & \textbf{0.417 {\tiny ±0.023}} \\
% \addlinespace
60
& 0.017 {\tiny ±0.004} & 0.202 {\tiny ±0.020}
& 0.103 {\tiny ±0.009} & 0.278 {\tiny ±0.021}
& 0.073 {\tiny ±0.009} & 0.359 {\tiny ±0.022}
& \textbf{0.217 {\tiny ±0.015}} & \textbf{0.415 {\tiny ±0.023}} \\
% \addlinespace
80
& 0.015 {\tiny ±0.005} & 0.184 {\tiny ±0.019}
& 0.068 {\tiny ±0.008} & 0.185 {\tiny ±0.017}
& 0.051 {\tiny ±0.006} & 0.343 {\tiny ±0.021}
& \textbf{0.176 {\tiny ±0.014}} & \textbf{0.422 {\tiny ±0.023}} \\
\bottomrule
\end{tabular}}
\caption{\textbf{Radiology report generation evaluation:} CheXmix (S1 + S2) achieves the best performance across masking percentages (best metrics in \textbf{bold}). We compute GREEN and CheXbert on a random sample of 1,000 images and report mean and standard deviation across three runs with different random seeds.}
\label{tab:rrg_main}
\end{table*}

\begin{figure*}[ht]
% \thisfloatpagestyle{empty}
\centerline{\includegraphics[width=1.0\textwidth]{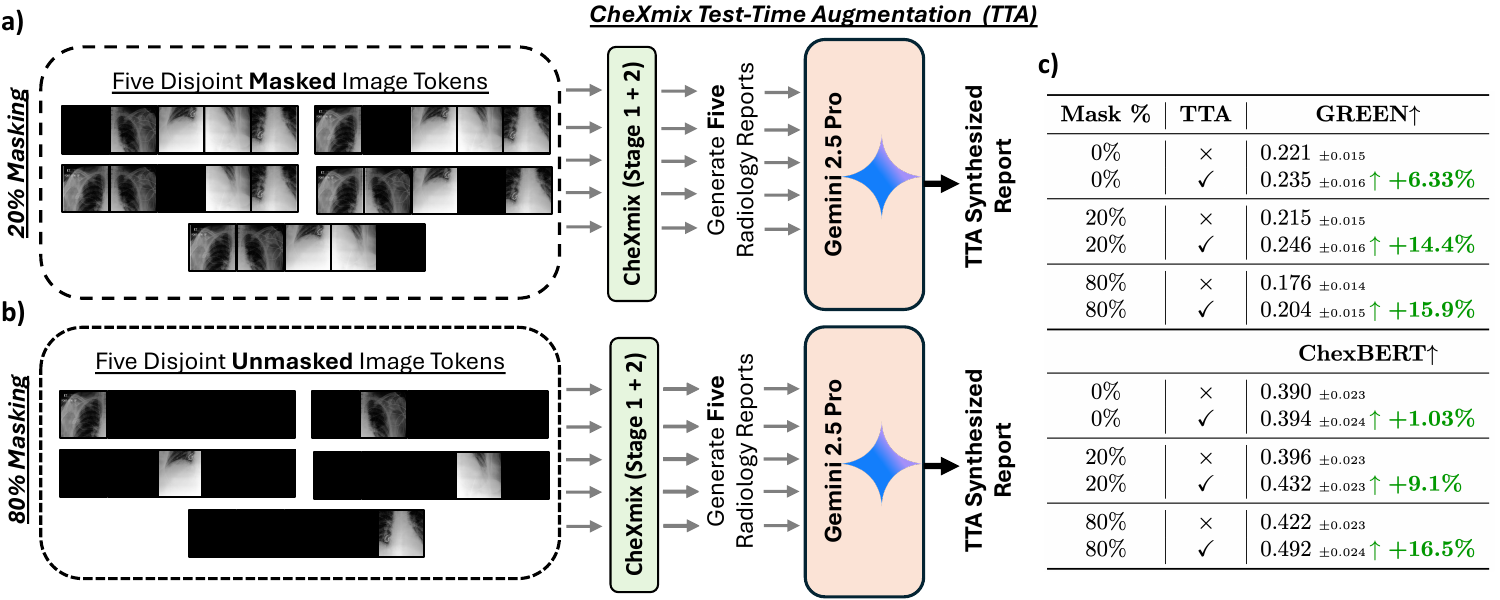}}
\caption{\textbf{Test-Time Augmentation with CheXmix.} Radiology report generation is improved by leveraging CheXmix in a test-time augmentation (TTA) setup. One image token sequence is converted into (a) five disjoint masked sequences at 20\% masking and (b) five disjoint unmasked sequences at 80\% masking. The masked indices are processed through CheXmix (Stage 1 + 2), and reports are synthesized using Gemini. TTA yields over 10\% improvement on GREEN and CheXbert at 20\% masking, and over 16\% improvement on CheXbert at 80\% masking.}
\label{fig:test_time_ensembling}
\end{figure*}

\subsection{Multimodal Retrieval}

We observe largely comparable retrieval performance (Top-8 and Top-16) across varying pool sizes (N = 32, 64, 128) between CheXmix (S1), CheXmix (S1 + S2), and CheXagent (SigLIP) (Table~\ref{tab:retrieval_comparison_percent_bold}). We evaluate both image-to-text (chest X-ray → radiology report) and text-to-image (radiology report → chest X-ray) recall across chunked pool sizes of 32, 64, and 128 over 2,048 test samples. For image-to-text retrieval, all models perform within 1\% of each other. For text-to-image retrieval, CheXagent slightly outperforms CheXmix; however, CheXmix (S1 + S2) remains within 2\% of CheXagent for Top-8 pools and within 1\% for Top-16 pools. Despite SigLIP being explicitly trained for retrieval, these results demonstrate strong image–text alignment in CheXmix’s multimodal generative embeddings.

\begin{table}[ht]
\centering
\resizebox{0.5\textwidth}{!}{
\begin{tabular}{c c c c c}
\hline
Top K & N & Model & I $\rightarrow$ T Recall (\%) & T $\rightarrow$ I Recall (\%) \\
\hline
\multirow{9}{*}{8} 
& \multirow{3}{*}{32} & CheXmix (S1)       & 24.9 [21.9, 28.1] & 22.3 [15.6, 29.5] \\
&                     & CheXmix (S1 + S2)       & \textbf{24.9 [21.9, 26.3]} & 23.0 [14.3, 28.1] \\
&                     & CheXagent (SigLIP) & 24.7 [15.6, 34.4] & \textbf{24.4 [15.6, 34.4]} \\
\cline{2-5}
& \multirow{3}{*}{64} & CheXmix (S1)       & 12.4 [10.6, 14.1] & 9.42 [4.34, 14.4] \\
&                     & CheXmix (S1 + S2)       & \textbf{12.6 [10.9, 14.1]} & 10.6 [7.11, 12.9] \\
&                     & CheXagent (SigLIP) & 12.5 [7.81, 17.2] & \textbf{12.6 [8.67, 17.2]} \\
\cline{2-5}
& \multirow{3}{*}{128} & CheXmix (S1)       & 6.25 [5.76, 6.74] & 4.59 [2.64, 5.96] \\
&                      & CheXmix (S1 + S2)       & 6.54 [5.47, 8.30] & 5.37 [3.91, 7.03] \\
&                      & CheXagent (SigLIP) & \textbf{6.88 [3.71, 10.6]} & \textbf{5.91 [3.12, 8.30]} \\
\hline
\multirow{9}{*}{16} 
& \multirow{3}{*}{32} & CheXmix (S1)       & \textbf{50.1 [46.9, 53.1]} & 48.8 [40.6, 57.6] \\
&                     & CheXmix (S1 + S2)       & 49.6 [46.9, 53.1] & 48.9 [40.6, 56.3] \\
&                     & CheXagent (SigLIP) & 49.7 [40.6, 60.7] & \textbf{49.5 [39.3, 59.4]} \\
\cline{2-5}
& \multirow{3}{*}{64} & CheXmix (S1)       & \textbf{25.0 [23.4, 26.6]} & 22.4 [17.2, 28.1] \\
&                     & CheXmix (S1 + S2)       & 25.0 [23.4, 26.6] & 22.4 [18.8, 26.9] \\
&                     & CheXagent (SigLIP) & 24.6 [16.8, 30.4] & \textbf{25.1 [20.0, 30.7]} \\
\cline{2-5}
& \multirow{3}{*}{128} & CheXmix (S1)       & 12.3 [10.9, 13.3] & 9.33 [7.03, 11.7] \\
&                      & CheXmix (S1 + S2)       & \textbf{12.5 [11.7, 13.3]} & 11.0 [9.38, 13.0] \\
&                      & CheXagent (SigLIP) & 12.4 [9.38, 15.3] & \textbf{11.7 [9.38, 14.6]} \\
\hline
\end{tabular}}
\caption{\textbf{Multimodal Retrieval.} We compare CheXmix and CheXagent on image–report retrieval, reporting mean Top-8 and Top-16 accuracy (\%) with 95\% confidence intervals. Best-performing values are highlighted in bold.}
\label{tab:retrieval_comparison_percent_bold}
\end{table}

\subsection{Ablations}

For additional pretraining ablations please refer to Section \ref{appendix:section_c2} of the supplementary material.

\paragraph{Causal Mask Ablation} 

We evaluate the effect of the causal mask (CM) and bidirectional attention in CheXmix’s pretraining stages through classification and report generation experiments (Table~\ref{tab:causal_mask_modeling}). For CheXmix (S1 + S2), models were pretrained with 50\% masking using either CM or bidirectional attention. Across masking ratios, CheXmix (S1 + S2) with CM generally performs better on classification and consistently outperforms bidirectional models in report generation. On the classification task, CheXmix (S1 + S2) with CM outperforms its bidirectional counterpart by 2.5\% across all masking percentages, whereas for S1, the bidirectional variant shows a 7.4\% improvement over S1 with CM. Additionally, CheXmix models with CM achieve a 41.0\% improvement on average in radiology report generation relative to their bidirectional counterparts. Compared to approaches such as Show-O and Transfusion~\citep{xie2024show, zhou2024transfusion}, which allow full attention across image tokens, we find that causal modeling of both image and text tokens is beneficial in the medical domain. Since CheXmix (S1 + S2) with CM achieves the strongest performance across tasks, all main-paper experiments use the CM models.

\begin{table}[h!]
\centering
\resizebox{0.5\textwidth}{!}{
\begin{tabular}{c|cccc}
\multicolumn{5}{l}{\textbf{a)\hspace{25mm}CheXpert Classification (AUROC)}} \\  
\midrule
Mask \% & S1 (B) & S1 (CM) & S1 + S2 (B) & S1 + S2 (CM) \\
\midrule
0  & 0.713 {\tiny ±0.000} & 0.664 {\tiny ±0.000} & \textbf{0.716} {\tiny ±0.000} & 0.712 {\tiny ±0.000} \\
20 & 0.702 {\tiny ±0.000} & 0.660 \tiny{$\pm$0.000}     & 0.684 {\tiny ±0.000} & \textbf{0.705} {\tiny ±0.000}     \\
40 & 0.678 {\tiny ±0.000} & 0.653 {\tiny ±0.001}     & 0.691 {\tiny ±0.000} & \textbf{0.702} {\tiny ±0.000} \\
60 & 0.667 {\tiny ±0.000} & 0.643 {\tiny ±0.000}     & 0.682 {\tiny ±0.000} & \textbf{0.689} {\tiny ±0.000} \\
80 & 0.591 {\tiny ±0.0017} & 0.624 {\tiny ±0.000}     & 0.632 {\tiny ±0.0000} & \textbf{0.656} {\tiny ±0.000} \\
\bottomrule
\addlinespace[1mm]
\multicolumn{5}{l}{\textbf{b)\hspace{25mm}Report Generation (GREEN metric)}} \\
\midrule
Mask \% & S1 (B) & S1 (CM) & S1 + S2 (B) & S1 + S2 (CM) \\
\midrule
0  & 0.172 {\tiny ±0.015} & 0.219 {\tiny ±0.015} & 0.140 {\tiny ±0.014} & \textbf{0.221 {\tiny ±0.015}} \\
20 & 0.153 {\tiny ±0.014} & 0.181 {\tiny ±0.013} & 0.141 {\tiny ±0.015} & \textbf{0.215 {\tiny ±0.015}} \\
40 & 0.112 {\tiny ±0.013} & 0.148 {\tiny ±0.012} & 0.132 {\tiny ±0.014} & \textbf{0.224 {\tiny ±0.015}} \\
60 & 0.071 {\tiny ±0.009} & 0.073 {\tiny ±0.009} & 0.129 {\tiny ±0.014} & \textbf{0.217 {\tiny ±0.015}} \\
80 & 0.061 {\tiny ±0.005} & 0.051 {\tiny ±0.006} & 0.104 {\tiny ±0.013} & \textbf{0.176 {\tiny ±0.014}} \\
\bottomrule
\end{tabular}
}
\caption{\textbf{CheXmix causal mask ablation.} We evaluate CheXmix pretrained with 50\% masking either using bidirectional attention (B) or a causal mask (CM) across classification and report generation tasks. Across masking ratios, CheXmix S1 + S2 with CM generally performs best for both tasks. Consequently, all main-paper CheXmix experiments use the causal mask.}
\label{tab:causal_mask_modeling}
\end{table}

\subsection{External Validation}

\paragraph{Extended Validation}
We evaluate CheXmix on two external datasets: ChestX-ray14 (classification) and ReXGradient (report generation). We find CheXmix consistently outperforms HealthGPT and other early-fusion models (Table ~\ref{tab:external_validation}). Moreover, CheXmix surpasses CheXagent (GREEN: {0.135\scriptsize$\pm$0.011}; CheXbert: {0.310\scriptsize$\pm$0.024}) on ReXGradient report generation, while CheXagent achieves an AUROC of {0.849\scriptsize$\pm$0.001} on ChestX-ray14.

\begin{table}[h!]
\centering
\resizebox{\linewidth}{!}{%
    \begin{minipage}{12cm} % Ample width to prevent text wrapping/squashing
        \centering
        \setlength{\tabcolsep}{4pt} % Comfortable padding between columns
        \textbf{a) CheXpert Cls. — AUROC (Table 1; No Masking)}
        \begin{tabular}{cccc}
            \toprule
            % & \multicolumn{2}{c}{\textbf{Generative Pretraining Objective}} & \textbf{Reference Max} \\
            % \cmidrule(r){1-3} \cmidrule(l){4-4}
            Chameleon & HealthGPT & M3AE & CheXmix (S1 + S2) \\
            % \cmidrule(r){1-3} \cmidrule(l){4-4}
            \midrule
            0.695 \tiny{$\pm$0.000} & 0.641 \scriptsize{$\pm$0.002} & 0.676 \tiny{$\pm$0.000} & \textbf{0.712 \scriptsize{$\pm$0.001}}  \\
            \bottomrule
        \end{tabular}

        \vspace{0.5em}

        % --- Table B ---
        \textbf{b) ChestX-ray14 Cls. — AUROC (External; No Masking)}
        \begin{tabular}{cccc}
            \toprule
            % \multicolumn{4}{c}{\textbf{Generative Pretraining Objective}} & \textbf{Reference Max} \\
            % \cmidrule(r){1-4} \cmidrule(l){5-5}
            Chameleon & HealthGPT & M3AE & CheXmix (S1+S2) \\
            \midrule
            % \cmidrule(r){1-4} \cmidrule(l){5-5}
            0.770 \scriptsize{$\pm$0.001} & 0.716 \scriptsize{$\pm$0.000} & 0.752 \scriptsize{$\pm$0.001} & \textbf{0.773 \scriptsize{$\pm$0.000}} \\
            \bottomrule
        \end{tabular}

        \vspace{0.5em}

        % --- Table C ---
        \textbf{c) ReXGradient Report Generation — Test Set (External)}        
        \begin{tabular}{cccccc}
            \toprule
            \multicolumn{2}{c}{Chameleon} & \multicolumn{2}{c}{HealthGPT}  & \multicolumn{2}{c}{CheXmix (S1 + S2)} \\
            \cmidrule(r){1-2} \cmidrule(lr){3-4} \cmidrule(l){5-6}
            GREEN$\uparrow$ & ChexBERT$\uparrow$ & GREEN$\uparrow$ & ChexBERT$\uparrow$ & GREEN$\uparrow$ & ChexBERT$\uparrow$ \\
            \midrule
            0.0287 \scriptsize{$\pm$0.001} & 0.139 \scriptsize{$\pm$0.017}
            & 0.216 \scriptsize{$\pm$0.017} & 0.239 \scriptsize{$\pm$0.021}
            % & 0.135 \scriptsize{$\pm$0.011} & 0.310 \scriptsize{$\pm$0.024} 
            & \textbf{0.217 \scriptsize{$\pm$0.016}} & \textbf{0.413 \scriptsize{$\pm$0.027}} \\
            \bottomrule
        \end{tabular}
        % \textbf{Table 7.} CheXmix outperforms other early-fusion models.
    \end{minipage}
}
\caption{\textbf{CheXmix External Validation.} CheXmix outperforms other early-fusion generative multimodal models across (a) CheXpert classification, (b) NIH ChestX-ray14 classification (external), and (c) ReXgradient report generation (external).}
\label{tab:external_validation}
\end{table}

\section{Discussion}

We leverage the complementary strengths of MAEs and MLLMs in a unified generative pretraining strategy for discriminative and generative medical image tasks. Unlike contrastive pretraining, CheXmix provides multimodal alignment while enabling fine-grained generative capabilities. By encoding image and text tokens into a shared vocabulary, our framework is highly scalable and flexible, supporting diverse downstream tasks from classification to report generation. Notably, CheXmix (S1 + S2) outperforms CheXagent by 8.6\% in AUROC on CheXpert classification at higher masking ratios and by 45.0\% on the GREEN metric on radiology report generation. CheXmix (S1 + S2) exhibits greater robustness to image corruption due to enriched representations from masked image–language pretraining. For CheXmix, a practical consideration is the increased computational cost of longer sequence lengths in transformer decoders. Our results suggest that unified early-fusion generative models could serve as a viable alternative to LLaVA-style models and offer a scalable solution for designing the next generation of medical FMs.

% Furthermore, CheXmix’s masked image–language pretraining strategy is compatible with transformer decoder architectures beyond RadPhi-2, highlighting its broader applicability.
% CheXmix (S1 + S2) exhibits greater robustness to image corruption, an important property for clinical settings where scans often contain artifacts or occlusions. 
\section{Acknowledgments}

A.K. completed this work during an internship at Oracle Health AI. A.K. is supported by graduate fellowship awards from the Knight-Hennessy Scholar program at Stanford University and the Tau Beta Pi Society. We would like to thank other members of Oracle Health AI for their support while developing our system and training our models, and 
Raefer Gabriel,
Sri Gadde,
Neil Hauge,
Samyak Jhaveri,
Mark Johnson,
Devashish Khatwani,
Ganesh Kumar,
Yuan-Fang Li,
Anit Sahu,
Amitabh Saikia,
Gyan Shankar,
Praphul Singh,
and
Vishal Vishnoi
for insightful feedback and discussions. We would also like to thank Christian Bluethgen at Stanford University for his expert review of model-generated reports during the review process.  This material is based on work supported by the Chameleon Research License, Copyright (c) Meta Platforms, Inc, All Rights Reserved. A.C. receives research support from NIH grants R01 HL167974, R01HL169345, R01 AR077604, R01 EB002524, R01 AR079431, P41 EB 027060, P50 HD118632; Advanced Research Projects Agency for Health (ARPA-H) Biomedical Data Fabric (BDF) and Chatbot Accuracy and Reliability Evaluation (CARE) programs (contracts AY2AX000045 and 1AYSAX0000024-01); and the Medical Imaging and Data Resource Center (MIDRC), which is funded by the National Institute of Biomedical Imaging and Bioengineering (NIBIB) under contract 75N92020C00021 and through ARPA-H.
{
    \small
    \bibliographystyle{ieeenat_fullname}
    \bibliography{main}
}


\onecolumn

%%% --- Add this block to create the supplementary header --- %%%
\begin{center}
    {\Large \bf CheXmix: Unified Generative Pretraining for Vision Language Models in Medical Imaging \par}
    \vspace{1em}
    {\Large \textbf{Supplementary Material} \par}
    \vspace{2em}
\end{center}
%%% --------------------------------------------------------- %%%

\appendix
\tableofcontents
\resumetocwriting
\setcounter{figure}{4} 
\setcounter{table}{6}
% \clearpage
% \setcounter{page}{1}
% \maketitlesupplementary
% \onecolumn
% \resumetocwriting

% \tableofcontents

\clearpage
\section{CheXmix Training Hyperparameters}
\label{appendix:section_a}

\begin{table*}[!ht] % Use table* for a full-width table in a two-column layout
    \centering % Center the table on the page
    \resizebox{0.6\textwidth}{!}{
    \begin{tabular}{l c c}
        \toprule % Top horizontal line
        \textbf{Configuration} &
        \textbf{\begin{tabular}{@{}c@{}}CheXmix (S1)\end{tabular}} &
        \textbf{\begin{tabular}{@{}c@{}}CheXmix (S2)\end{tabular}} \\
        \midrule % Middle horizontal line separating header from body
        LLM Init.                & RadPhi-2     & from Stage 1    \\
        Image Resolution         & $512^2$   & $512^2$           \\
        Masking Percentage       & - & 50\% \\
        LLM max sequence length      & 1300     & 1300 \\
        Optimizer                & \multicolumn{2}{c}{AdamW}                                                 \\
        Optimizer hyperparameter & \multicolumn{2}{c}{$\beta_1 = 0.9$, $\beta_2 = 0.98$, \textit{eps} $= 1\text{e}-6$} \\
        Peak learning rate       & 1e-5      & 1e-5 \\
        Learning rate schedule   & \multicolumn{2}{c}{cosine decay}                                          \\
        Weight decay             & 0.1     & 0.1               \\
        Gradient clip            & 1.0         & 1.0 \\
        Training steps          & 703,671         & 513,993 \\
        Global batch size        & 8     & 8 \\
        Gradient Acc.            & 4         & 4                 \\
        Numerical precision      & bfloat16    & bfloat16 \\
        \bottomrule % Bottom horizontal line
    \end{tabular}}
    \caption{Pretraining hyperparameters for CheXmix S1 and S2 models.}
    \label{suppl_tab:training_configs}
\end{table*}

\FloatBarrier

\section{Extended Methods}\label{appendix:extended_methods}

\subsection{Model Pretraining}
\label{appendix:section_b1}

\paragraph{Additional Staged Pretraining Explanation:}

We provide a simplified explanation below to clarify our training objectives and naming conventions. Both stages use a next-token prediction loss, but differ in whether input tokens are masked. We provide detailed hyperparameters for pretraining in Table ~\ref{suppl_tab:training_configs}. Our models are pretrained on both chest X-ray image tokens and radiology report text tokens using a two-stage approach, detailed below:

\begin{enumerate}
    \item \textbf{Stage 1: Standard Autoregressive Pretraining:}  
    This is the conventional autoregressive language modeling objective.  
    Given an input sequence $(t_1, t_2, \ldots, t_N)$, the model predicts the next token at each position, and the cross-entropy loss is summed over the entire sequence.

    \item \textbf{Stage 2: Masked Image-Language Pretraining:}  
    In Stage~2, we introduce an autoregressive masked-token prediction loss that combines ideas from autoregressive modeling and masked autoencoders.  
    Similar to Stage~1, the model predicts the next token at each position; however, we randomly replace 50\% of the input image and text tokens with a special \texttt{[MASK]} token, and we compute the loss \emph{only} for output tokens that immediately follow a masked input token (Figure~2b).  
    In effect, the model must reconstruct information missing from its input by leveraging corrupted context.

    \vspace{1mm}
    For example, if the input sequence is $(t_1, t_2, t_3)$ and we randomly mask position 2, we obtain $(t_1, \texttt{[MASK]}, t_3)$.  
    The model then predicts $(p_2, p_3, p_4)$.  
    We ignore the losses for $p_2$ (predicted after $t_1$) and $p_4$ (predicted after $t_3$), and compute the loss only for $p_3$, which is predicted after the masked token and is compared against the ground-truth $t_3$.  
    This design encourages the model to accurately predict tokens following masked inputs, and our evaluations indicate strong representation quality and improved robustness to masked input under this strategy.
\end{enumerate}

\subsection{Evaluation Methods}
\label{appendix:section_b2}

We present details of our evaluation pipeline. In general, we conduct a rigorous analysis of representational quality through both discriminative and generative tasks. Our evaluation suite includes CheXpert embedding findings classification, image inpainting, radiology report generation, multimodal retrieval, test-time augmentation for report generation, and several ablation studies. We first assess the discriminative capability of CheXmix's embeddings by evaluating pretrained representations on the CheXpert dataset and comparing them to relevant general-domain and medical-specific baselines. Next, we evaluate CheXmix's generative capabilities through image inpainting and radiology report generation, using a test set composed of images and reports from five datasets (Section 3.1; Main Paper). 

For both classification and generation tasks, we examine model performance across multiple masking percentages (20\%, 40\%, 60\%, and 80\%) to highlight CheXmix's fine-grained representational capacity. The rationale for masking during evaluation is to provide a general assessment of each model's representational robustness to partial or occluded inputs; in real-world chest radiographs, regions of interest may be obscured by medical devices (e.g., pacemakers, ECG leads)~\cite{mathew2019chest}, imaging artifacts~\cite{shetty2011computed}, or overlapping anatomy~\cite{samei2003subtle}, and a robust model should leverage global context to make accurate predictions despite missing information.  We further demonstrate improvements in report quality using CheXmix's test-time augmentation strategy, which does not require additional training, and we evaluate the impact of causal masking versus bidirectional attention during pretraining, as well as the effect of different masking ratios.

\paragraph{CheXpert Findings Classification:} \label{text:chexpert_findings_cls} We evaluate pretrained representations on the CheXpert dataset using a multi-head masked linear probe classification task over 14 findings: \textit{Enlarged Cardiomediastinum, Cardiomegaly, Lung Opacity, Lung Lesion, Edema, Consolidation, Pneumonia, Atelectasis, Pneumothorax, Pleural Effusion, Pleural Other, Fracture, Support Devices, and No Finding}. The pretrained embedding dimensions are as follows: Chameleon (4096), HealthGPT (1024), MAE (384), M3AE (768), CheXagent (2560), CheXmix (Stage 1 and 2, 2560). For CheXmix, we tokenize images using the VQ-GAN tokenizer developed with Chameleon \citep{team2024chameleon}. We first process the embeddings for the images on the CheXpert dataset and take the mean of the embeddings across all image patches to get a single embedding vector for the image. Averaging token embeddings has been shown to result in better performance than probing at other token positions~\citep{zhang2024visually}.

In generative models, unlike discriminative models, the information relevant for classification is not necessarily concentrated at the final layer. Therefore, we extract embeddings from every layer of each model. Specifically, we consider Chameleon (32 layers), HealthGPT (24 layers), MAE (13 layers), M3AE (13 layers), CheXagent (26 layers), and CheXmix (32 layers). For each model, we select the layer that achieves the highest AUROC on the validation set and report the corresponding AUROC and AUPRC metrics on the test set. The middle
layers yield the best embeddings for generatively pretrained
models (Chameleon, CheXmix), whereas the final layers perform best for vision encoder models (M3AE, CheXagent) (Table ~\ref{tab:model_best_layers}).

 For each model, we train 14 linear probes with bias terms, optimized using AdamW (without weight decay) for 100 epochs with batch size 8 and gradient accumulation of 8. Training is conducted with an initial learning rate of $1\times 10^{-5}$, cosine learning rate decay, and without mixed-precision. To examine robustness under different levels of supervision, we apply masking ratios of 20\%, 40\%, 60\%, and 80\% during training. For the generatively pretrained models (Chameleon, MAE, M3AE, CheXmix), masking is applied at the token level, whereas for CheXagent and HealthGPT, which inputs image patches directly into the transformer, masking is applied at the image level. CheXpert labels are provided as $-1$ (uncertain/missing), 0, 1, or empty; we treat empty cells as $-1$ and mask the loss for labels equal to $-1$. We process the CheXpert training data and generate train/validation/test splits of 22,342 / 234 / 667, respectively. Performance is measured using AUROC and AUPRC and all reported metrics metrics are averaged over three random seeds. The goal of this task is to isolate and evaluate the quality of pretrained representations.

\begin{table}[ht]
\centering
\begin{tabular}{lccccc}
\hline
\textbf{Model} & \multicolumn{5}{c}{\textbf{Masking Percentage (\%)}} \\
\cline{2-6}
& \textbf{0\%} & \textbf{20\%} & \textbf{40\%} & \textbf{60\%} & \textbf{80\%} \\
\hline
Chameleon & 4 & 4 & 7 & 18 & 1 \\
HealthGPT & 23 & 17 & 23 & 18 & 17 \\
MAE & 11 & 9 & 8 & 9 & 11 \\
M3AE & Trans. 10 & Trans. 8 & Trans. 8 & Trans. 9 & Trans. 10 \\
CheXmix (S1) & 8 & 10 & 9 & 8 & 18 \\
CheXmix (S1 + S2) & 8 & 8 & 8 & 8 & 8 \\
CheXmix (S1; B) & 14 & 15 & 12 & 12 & 7 \\
CheXmix (S2; B) & 10 & 3 & 5 & 5 & 32 \\
CheXagent & Encoder 23 & Encoder 16 & Encoder 10 & Encoder 10 & Encoder 8 \\
\hline
\end{tabular}
\caption{Best-Performing Layers Across Masking Percentages. Models vary in their number of layers (e.g., CheXmix has 33). For generatively pretrained models (Chameleon, CheXmix), intermediate layers produce the strongest embeddings, whereas for vision encoder models (M3AE, CheXagent), the final layers perform best.}
\label{tab:model_best_layers}
\end{table}

\paragraph{Image Inpainting:} We evaluate inpainting performance by reconstructing masked regions of images and assessing quality using similarity metrics including PSNR, MS-SSIM, and FID-Inception, implemented via the \texttt{torchmetrics} library. For evaluation, we randomly sample 5,000 images from the validation split of our pretraining dataset, which consists of five different datasets, and tokenize them with the VQ-GAN tokenizer \citep{team2024chameleon}. We experiment with masking ratios ranging from 10–90\%. For the CheXmix (Stage 1 and 2) models, the designated mask token is 58,560. At each masking ratio, we randomly generate indices to mask within the token sequence and feed the partially masked sequence through the model, predicting replacements by selecting the token with the highest probability score. We then decode the predicted tokens back into image space. As a baseline, we also measure reconstruction quality by tokenizing an image, applying masking, and then directly decoding the tokens back into image space using the VQ-GAN decoder without generative modeling. The goal of this task is to measure the model’s ability to reconstruct high-fidelity visual details from incomplete observations.

\paragraph{Radiology Report Generation:} \label{suppl_para:rrg} We evaluate radiology report generation by providing images with varying levels of masked input and generating the corresponding reports, focusing on both the findings and impression sections. Performance is assessed using domain-specific metrics including GREEN~\citep{ostmeier2024green}, CheXbert~\cite{smit2020CheXbert}, RadGraph-F1~\cite{jain2021radgraph}, and BERTScore~\cite{zhang2019bertscore}. For evaluation, we randomly sample 1000 image–text pairs from the validation split of our pretraining dataset, which consists of five different datasets, and tokenize the images using a VQ-GAN tokenizer. We experiment with masking ratios ranging from 10–90\%. At each ratio, we randomly generate indices to mask within the token sequence and feed the partially masked sequence and $T_S$ (text-start token) through the model to generate the associated report. To ensure fair comparison across samples, we set the maximum token limit for generated reports equal to the length of the original input report. As a baseline, we also evaluate the Chameleon model, which is masked at the token level. For Chameleon, we provide the prompt: \texttt{Generate a findings and impression section for this chest X-ray image. Include the 'findings' and 'impressions' tag in the report. Do not list the findings and impressions separately; instead, present them in one continuous section.} For CheXagent, we provide the prompt: \texttt{Generate the findings and impression section.}

\paragraph{Radiology Report Generation (Test-Time Augmentation):}
To illustrate a practical use case of masked learning and better motivate CheXmix (S2) pretraining, we process over 1,000 generated reports using a Test-Time Augmentation (TTA) strategy. Specifically, we introduce a disjoint masking protocol that generates multiple, distinct masked versions of each image, allowing the model to generate radiology reports from these masked image tokens. 

Let $\mathcal{I} = \{1, \dots, N\}$ be the set of all $N=1024$ image token indices. We partition $\mathcal{I}$ into $K=5$ mutually disjoint subsets $\mathcal{S}_1, \dots, \mathcal{S}_K$, such that every token belongs to exactly one subset:
\begin{equation}
    \bigcup_{k=1}^{K} \mathcal{S}_k = \mathcal{I} 
    \quad \text{and} \quad 
    \mathcal{S}_k \cap \mathcal{S}_j = \emptyset, \quad \forall k \neq j.
\end{equation}
This partition ensures that the subsets $\mathcal{S}_k$ are unique and non-overlapping. Using this partition, we define the set of \textit{visible} tokens, denoted as $\mathcal{V}_k$, for the $k$-th input variation under two distinct settings:
\begin{enumerate}[leftmargin=*]
    \item \textbf{20\% Masking (Disjoint Masks):} In this setting, we mask the tokens in $\mathcal{S}_k$ while keeping the remainder visible. Formally, the visible set is defined as the complement:
    \begin{equation}
        \mathcal{V}_k = \mathcal{I} \setminus \mathcal{S}_k.
    \end{equation}
    Across the $K$ variations, the masked region shifts such that a different 20\% of the image is hidden in each pass, allowing the model to generate radiology reports from complementary subsets of visual information.

    \item \textbf{80\% Masking (Disjoint Views):} In this setting, we keep \textit{only} the tokens in $\mathcal{S}_k$ visible, masking the remaining $80\%$. The visible set is:
    \begin{equation}
        \mathcal{V}_k = \mathcal{S}_k.
    \end{equation}
    Here, the unmasked regions are disjoint across the $K$ variations. This forces the model to attend to distinct visual information in each pass for generating radiology reports. 
\end{enumerate}

Gemini 2.5 Pro~\cite{comanici2025gemini} (\texttt{gemini-2.5-pro}) then consolidates the unique characteristics from the five radiology reports, generated from the 20\% and 80\% masked inputs, into a single synthesized report. We evaluate the performance of this TTA strategy using the GREEN and CheXbert metrics, producing reports as described in Section~\ref{suppl_para:rrg}. The primary intuition behind TTA is that sampling the model multiple times with different masked and unmasked image inputs captures variations in its predictions, thereby probing the model’s epistemic uncertainty. An example of the Gemini prompt used in this process is provided in Section~\ref{suppl_sec:TTA_prompt}.

\paragraph{Multimodal Retrieval:} We evaluate image–text (chest X-ray → radiology report) and text–image (radiology report → chest X-ray) retrieval performance for CheXmix (S1), CheXmix (S1+S2), and CheXagent (SigLIP). Retrieval is assessed using Top-8 and Top-16 accuracy across chunked pool sizes of 32, 64, and 128 over 2,048 test samples. For each image–text pair, we compute cosine similarity between the corresponding embeddings and rank them accordingly. Chest X-rays are 512×512 images tokenized into 1,024 discrete tokens and processed through the unified transformer decoder for CheXmix, and patch-wise (32×32) through the vision transformer for SigLIP. Radiology reports, composed of the Findings and Impression sections, are encoded using the CheXmix unified transformer decoder or the SigLIP text encoder. The resulting embeddings are 2,056-dimensional for both image and text in CheXmix, and 1,024-dimensional for both modalities in SigLIP. For all models, we average token embeddings to produce a single vector representation per modality for retrieval.

\paragraph{Pretraining Ablation (Masking Ratio):}
We evaluate the effect of masking ratio during CheXmix (S2) pretraining by training models with four ratios: 25\%, 50\%, 75\%, and 90\%. These values are motivated by medical-domain literature~\cite{xing2023self}, which shows that moderate masking ratios (e.g., 40\%) can yield strong representations for chest X-rays, and the seminal masked autoencoder work~\cite{he2022masked}, which recommends higher ratios such as 75\% or 90\% to reduce spatial redundancy. We pretrain all models for 100K steps and then assess their downstream performance on the CheXpert embedding–based findings classification task. For each masking ratio, we compute AUROC and AUPRC by selecting the best-performing layer among the 32 layers of the transformer decoder using the validation set, and we report the corresponding performance on the test set.

\paragraph{Pretraining Ablation (Causal Mask):}
We evaluate pretraining with and without applying a causal mask to the image tokens at both CheXmix S1 and S2 pretraining stages. Recent unified generative models in the general domain, such as Show-O~\cite{xie2024show} and Transfusion~\cite{zhou2024transfusion}, report strong performance when images are given full bidirectional attention while text tokens remain causally masked. This design reflects the intuition behind vision transformers, where bidirectional attention is appropriate because the causal ordering of image patches is not semantically meaningful. In contrast, other models such as Chameleon~\cite{team2024chameleon} maintain causal masking for both modalities.

In our approach, we enable full bidirectional attention over the 1,024 image tokens while keeping the text tokens strictly causal. We construct the attention mask by first initializing a standard causal mask $A \in \mathbb{R}^{N \times N}$  for the entire sequence of $N = N_{\text{img}} + N_{\text{text}}$ tokens. To convert the image portion to full attention, we overwrite the corresponding block of the attention matrix with zeros. Specifically, for image token indices $i, j \in \{1, \dots, N_{\text{img}}\}$, we update the mask such that $A_{ij} = 0$, ensuring bidirectional attention among image tokens while preserving causal masking for all positions involving text tokens. For detailed hyperparameters and training configurations, please refer to Table ~\ref{suppl_tab:training_configs_no_cm}.

\begin{table*}[!ht] % Use table* for a full-width table in a two-column layout
    \centering % Center the table on the page
    \resizebox{0.6\textwidth}{!}{
    \begin{tabular}{l c c}
        \toprule % Top horizontal line
        \textbf{Configuration} &
        \textbf{\begin{tabular}{@{}c@{}}CheXmix (S1; B)\end{tabular}} &
        \textbf{\begin{tabular}{@{}c@{}}CheXmix (S2; B)\end{tabular}} \\
        \midrule % Middle horizontal line separating header from body
        LLM Init.                & RadPhi-2     & from Stage 1    \\
        Image Resolution         & $512^2$   & $512^2$           \\
        Image Attention          & \multicolumn{2}{c}{Bidirectional (B)} \\
        Text Attention           & \multicolumn{2}{c}{Causal Mask (CM)} \\
        Masking Percentage       & - & 50\% \\
        LLM max sequence length      & 1300     & 1300 \\
        Optimizer                & \multicolumn{2}{c}{AdamW}                                                 \\
        Optimizer hyperparameter & \multicolumn{2}{c}{$\beta_1 = 0.9$, $\beta_2 = 0.98$, \textit{eps} $= 1\text{e}-6$} \\
        Peak learning rate       & 1e-5      & 1e-5 \\
        Learning rate schedule   & \multicolumn{2}{c}{cosine decay}                                          \\
        Weight decay             & 0.1     & 0.1               \\
        Gradient clip            & 1.0         & 1.0 \\
        Training steps          &  683,000     &  1,353,000 \\
        Num. of GPUs (A100)     & 8 & 4 \\
        Global batch size        & 16     & 8 \\
        Gradient Acc.            & 4         & 4                 \\
        Numerical precision      & bfloat16    & bfloat16 \\
        \bottomrule % Bottom horizontal line
    \end{tabular}}
    \caption{Pretraining hyperparameters for CheXmix S1 and S2 models with bidrectional attention for the image tokens.}
    \label{suppl_tab:training_configs_no_cm}
\end{table*}

\paragraph{NIH ChestX-ray14 (External Evaluation):} We evaluate pretrained representations on the NIH ChestX-ray14 dataset using a multi-head masked linear probe classification task across 14 findings: atelectasis, cardiomegaly, pleural effusion, infiltration, lung mass, lung nodule, pneumonia, pneumothorax, consolidation, edema, emphysema, fibrosis, pleural thickening, and hernia. We preprocess the dataset and create train/validation/test splits of 19,621 / 1,121 / 2,242, respectively. The embedding dimensions of the pretrained models are as follows: Chameleon (4096), HealthGPT (1024), M3AE (768), CheXagent (2560), and CheXmix (S1 + S2; 2560). We train linear probes following the “CheXpert Findings Classification” protocol (see~\ref{text:chexpert_findings_cls}), without evaluating across masking percentages. Performance is reported using AUROC, averaged over three random seeds. The best-performing layers used for evaluation are: Chameleon (layer 16), SigLIP (encoder\_23\_mean), M3AE (final layer), CheXmix (S1 + S2; layer 22), and HealthGPT (encoder\_23\_mean).

\paragraph{ReXgradient-160K (External Evaluation):} Similar to the "Radiology Report Generation" protocol (see ~\ref{suppl_para:rrg}), we evaluate radiology report generation by providing images (no masking) and generating the corresponding reports, focusing on both the findings and impression section. We select 500 random samples from the test set of ReXgradient-160K, specifically chest x-ray and radiology report samples. We evaluate these metrics over three seeds.

\FloatBarrier

\subsection{Evaluation Metrics}

\paragraph{CheXpert Embedding Findings Classification:} 
We evaluate the discriminative capability of CheXmix and baseline embeddings on the CheXpert findings classification task using AUROC and AUPRC. For each model, we select the probe (layer) that achieves the highest AUROC on the validation set and report its corresponding performance on the test set.

\begin{enumerate}
    \item \textbf{Area under the receiver operating characteristic curve
(AUROC):} AUROC evaluates a binary classification model’s ability to differentiate between positive and negative cases across all possible decision thresholds. It is computed as the area under the curve obtained by plotting the True Positive Rate (TPR, or sensitivity) against the False Positive Rate (FPR, equal to 1–specificity). An AUROC of 0.5 indicates performance equivalent to random chance, while a value of 1.0 denotes perfect discrimination.
    
    \item \textbf{Area under the precision-recall curve (AUPRC):} AUPRC measures a model’s ability to correctly identify positive cases across different decision thresholds, with particular emphasis on performance when the dataset is imbalanced. It is computed as the area under the curve obtained by plotting Precision (positive predictive value) against Recall (sensitivity). AUPRC emphasizes the model’s ability to detect the positive class, making it especially informative when positive cases are rare, as is often the case for many abnormalities. A higher AUPRC indicates that the model maintains strong precision even at high recall levels, demonstrating its capacity to correctly identify true positives while minimizing false positives.
\end{enumerate}

\paragraph{Image Inpainting:} We evaluate the inpainting performance of CheXmix and baseline models on PSNR, MS-SSIM, and FID.

\begin{enumerate}
    \item \textbf{Peak Signal-to-Noise Ratio (PSNR):} PSNR quantifies pixel-level reconstruction fidelity by measuring the ratio between the square of the maximum possible pixel value and the Mean Squared Error (MSE) between the ground truth and inpainted image, expressed in decibels (dB). A higher PSNR indicates that the generated image is numerically closer to the original in terms of pixel intensity values. 
    
    \item \textbf{Multi-Scale Structural Similarity Index Measure (MS-SSIM):} MS-SSIM evaluates the perceptual quality of the reconstruction by analyzing structural similarity across multiple scales and resolutions, capturing both global patterns and fine-grained local details. Ranging from 0 to 1, a higher MS-SSIM indicates that the model has preserved structural integrity, edge definition, and anatomical patterns, which are relevant for chest X-ray interpretation.

    \item \textbf{Fréchet Inception Distance (FID):} FID assesses the perceptual realism and diversity of generated images by measuring the distance between feature distributions of real and inpainted images in the embedding space of a pre-trained deep neural network (Inception-v3). Unlike PSNR and MS-SSIM, which rely on direct pixel-level comparisons with ground truth, FID evaluates embedding-level similarity by assessing whether the generated distribution matches the statistical and semantic properties of real data. A lower FID score indicates that the inpainted regions exhibit visual features consistent with the original chest X-rays, suggesting high perceptual quality.
\end{enumerate}

\paragraph{Radiology Report Generation:} We compute GREEN, CheXbert, RadGraph, and BERTScore to evaluate our generate radiology report across CheXmix and baselines.

\begin{enumerate}
    \item \textbf{Generative Radiology Report Evaluation and Error Notation (GREEN)~\cite{ostmeier2024green}:} GREEN is a clinically aligned metric that evaluates radiology report quality by identifying and explaining clinically significant errors. Unlike standard metrics such as BLEU or ROUGE, it leverages large language models to detect discrepancies and provides a quantitative score. A higher GREEN score indicates a report that is accurate, interpretable, and closely aligned with expert assessments, making it useful for improving automated radiology reporting.
    \item \textbf{CheXbert~\cite{smit2020CheXbert}:} CheXbert evaluates the clinical accuracy of generated reports by treating evaluation as a multi-label classification task. It uses a BERT-based labeler to extract the presence, absence, or uncertainty of 14 clinical observations (e.g., Pneumonia, Cardiomegaly, No Finding) from both the generated and reference reports. We report the weighted F1 score between the two label sets, which quantifies the model's ability to correctly identify clinical findings regardless of specific phrasing.
    \item \textbf{RadGraph~\cite{jain2021radgraph}:} RadGraph assesses the factual and structural completeness of reports by parsing them into clinical knowledge graphs containing entities (e.g., anatomical structures, observations, pathologies) and relations (e.g., "located at," "suggestive of"). By computing the F1 score based on the overlap of entities and relations between the generated and reference graphs, this metric rewards models that correctly capture clinical dependencies and anatomical relationships, rather than isolated keywords.
    \item \textbf{BERTScore~\cite{zhang2019bertscore}:} BERTScore evaluates the semantic similarity between generated and reference reports using token embeddings from a pre-trained language model. Unlike traditional metrics such as BLEU that rely on exact word matching, BERTScore computes cosine similarity between token representations, enabling it to recognize synonyms and paraphrases. This provides a measure of how well the overall meaning of the report are preserved
\end{enumerate}

\paragraph{Multimodal Retrieval:} Given a set of image–report pairs, we evaluate image-to-report retrieval using Recall@8 and Recall@16, which quantifies the proportion of test samples for which the correct report is retrieved among the top-8 or top-16 results, respectively. Retrieval is performed by computing the cosine similarity between image and text embeddings. Higher recall values indicate that the learned embedding space more effectively aligns visual and textual representations, enabling relevant image–report pairs to be retrieved more reliably.

\subsection{Baseline Justification} \label{appendix:section_b4}
We selected these baselines to provide a comprehensive comparison across both general-domain and medical-specific models.

\begin{enumerate}
    \item \textbf{CheXpert Findings Classification:} We isolate and evaluate the quality of pretrained image embeddings from each model using linear probes on the 14 CheXpert findings.
    \begin{enumerate}
        \item \textbf{Chameleon~\cite{team2024chameleon}:} A 7B-parameter general-domain multimodal generative model trained on 4.2T image and text tokens. Included as the most comparable method to our generative pretraining approach, since it unifies images and text as tokens and trains autoregressively.
        \item \textbf{HealthGPT~\cite{lin2025healthgptmedicallargevisionlanguage}}: A medical-specific early-fusion multimodal generative model that integrates clinical images and text for comprehension and generation.
        \item \textbf{RadPhi-2~\cite{chexagent-2024}:} Text-only model pretrained on 2.7T tokens of medical text, including radiology reports; serves as a token prediction baseline without visual context.
        \item \textbf{Masked Autoencoder (MAE)~\cite{he2022masked}:} Widely used masked image modeling baseline capturing strong visual representations; benchmarks a vision-only generative pretraining approach robust to image masking.
        \item \textbf{M3AE~\cite{chen2022multi}:} Multimodal masked autoencoder trained on chest X-rays and radiology reports, using a multimodal generative masking objective.
        \item \textbf{CheXmix (S1):} Stage 1 model trained jointly on chest X-rays and radiology reports to assess the advantage of unified generative pretraining over text-only modeling.
        \item \textbf{CheXmix (S1 + S2):} Stage 2 model building upon Stage 1 by introducing multimodal masked token prediction, allowing analysis of how masking improves classification performance.
        \item \textbf{CheXagent~\cite{chexagent-2024}:} State-of-the-art multimodal large language model (MLLM) for chest X-rays, included as an upper-bound domain-specific baseline. We use the SigLIP image encoder pretrained on over 8 million chest X-rays.
    \end{enumerate}

    \item \textbf{Image Inpainting:} Reconstruct full images from masked image tokens.
    \begin{enumerate}
        \item \textbf{VQ-GAN~\cite{team2024chameleon}:} Uses the Chameleon VQ-GAN image tokenizer to encode and decode masked image tokens; serves as a baseline for image reconstruction without full inpainting.
        \item \textbf{RadPhi-2~\cite{chexagent-2024}:} Text-only model; serves as a random token prediction baseline illustrating performance without visual information.
        \item \textbf{CheXmix (S1):} Stage 1 model trained autoregressively on chest X-rays and radiology reports; establishes a baseline for reconstructing masked regions without explicit masked pretraining.
        \item \textbf{CheXmix (S1 + S2):} Stage 2 model trained in a masked autoregressive manner; evaluates the impact of masked pretraining on image inpainting performance.
    \end{enumerate}

    \item \textbf{Radiology Report Generation:} Generate radiology reports from chest X-rays.
    \begin{enumerate}
        \item \textbf{Chameleon~\cite{team2024chameleon}:} Evaluates the transferability of general-domain pretraining to radiology report generation.
        \item \textbf{RadPhi-2~\cite{chexagent-2024}:} Text-only model serving as a token prediction baseline without visual context for radiology reports.
        \item \textbf{CheXmix (S1):} Stage 1 model trained autoregressively; generates reports by first inputting a chest X-ray image.
        \item \textbf{CheXmix (S1 + S2):} Stage 2 model trained in a masked autoregressive manner; evaluates the ability to generate reports from image tokens.
        \item \textbf{CheXagent~\cite{chexagent-2024}:} State-of-the-art MLLM for radiology report generation from chest X-rays.
    \end{enumerate} 
\end{enumerate}

\section{Extended Quantitative Results}\label{appendix:extended_results}

\subsection{Extended Main Paper Results} \label{appendix:section_c1}

\begin{table*}[htbp]
\centering
\small
\resizebox{\textwidth}{!}{%
\begin{tabular}{c|ccccccc}
\toprule
\multicolumn{7}{c|}{\textbf{Generative Pretraining Objective}} & \multicolumn{1}{c}{\textbf{Reference Maximum}} \\
\midrule
\textbf{Masking \%} & \textbf{Chameleon} & \textbf{HealthGPT} & \textbf{MAE} & \textbf{M3AE} & \textbf{CheXmix (S1)} & \textbf{CheXmix (S1 + S2)} & \textbf{CheXagent} \\
\midrule
\textbf{0\%}  & 0.361 \tiny{$\pm$0.000} & 0.319 \tiny{$\pm$0.000} & 0.333 \tiny{$\pm$0.001} & 0.359 \tiny{$\pm$0.000} & 0.333 \tiny{$\pm$0.000} & \underline{0.377 \tiny{$\pm$0.000}} & \textbf{0.520 \tiny{$\pm$0.003}} \\
\textbf{20\%} & 0.344 \tiny{$\pm$0.001} & 0.301 \tiny{$\pm$0.000} & 0.333 \tiny{$\pm$0.002} & \underline{0.350 \tiny{$\pm$0.001}} & 0.339 \tiny{$\pm$0.001} & 0.344 \tiny{$\pm$0.001} & \textbf{0.404 \tiny{$\pm$0.001}} \\
\textbf{40\%} & 0.325 \tiny{$\pm$0.000} & 0.247 \tiny{$\pm$0.001} & 0.330 \tiny{$\pm$0.004} & 0.351 \tiny{$\pm$0.000} & 0.331 \tiny{$\pm$0.001} & \textbf{0.368 \tiny{$\pm$0.001}} & 0.322 \tiny{$\pm$0.000} \\
\textbf{60\%} & 0.296 \tiny{$\pm$0.000} & 0.243 \tiny{$\pm$0.003} & 0.332 \tiny{$\pm$0.003} & 0.344 \tiny{$\pm$0.001} & 0.322 \tiny{$\pm$0.000} & \textbf{0.358 \tiny{$\pm$0.002}} & 0.283 \tiny{$\pm$0.001} \\
\textbf{80\%} & 0.256 \tiny{$\pm$0.000} & 0.217 \tiny{$\pm$0.001} & 0.310 \tiny{$\pm$0.003} & 0.327 \tiny{$\pm$0.000} & 0.281 \tiny{$\pm$0.001} & \textbf{0.337 \tiny{$\pm$0.001}} & 0.224 \tiny{$\pm$0.001} \\
\bottomrule
\end{tabular}%
}
\caption{\textbf{CheXpert Embedding Findings Classification.} CheXmix (S1 + S2) demonstrates superior AUPRC performance at higher masking percentages compared to other generative baselines. \textbf{Bold} indicates the best-performing model for each masking percentage, while \underline{underline} marks the best-performing generative model. AUPRC (mean~$\pm$~std) is reported across three random seeds.}
\label{tab:chexpert_auprc_std}
\end{table*}

\begin{table*}[htbp]
\centering
\small
\resizebox{\textwidth}{!}{%
\begin{tabular}{c|ccc|ccc|ccc|ccc}
\toprule
\multirow{2}{*}{\textbf{Masking \%}} & \multicolumn{3}{c|}{\textbf{VQ-GAN}} & \multicolumn{3}{c|}{\textbf{RadPhi-2}} & \multicolumn{3}{c|}{\textbf{CheXmix (S1)}} & \multicolumn{3}{c}{\textbf{CheXmix (S1 + S2)}} \\
\cmidrule(lr){2-4} \cmidrule(lr){5-7} \cmidrule(lr){8-10} \cmidrule(lr){11-13}
 & \textbf{PSNR}$\uparrow$ & \textbf{MS-SSIM}$\uparrow$ & \textbf{FID}$\downarrow$ & \textbf{PSNR}$\uparrow$ & \textbf{MS-SSIM}$\uparrow$ & \textbf{FID}$\downarrow$ & \textbf{PSNR}$\uparrow$ & \textbf{MS-SSIM}$\uparrow$ & \textbf{FID}$\downarrow$ & \textbf{PSNR}$\uparrow$ & \textbf{MS-SSIM}$\uparrow$ & \textbf{FID}$\downarrow$ \\
\midrule
\textbf{10} & 15.00{\scriptsize$\pm$0.01} & 0.671{\scriptsize$\pm$0.00} & 236.5{\scriptsize$\pm$0.75} & 22.55{\scriptsize$\pm$0.04} & 0.856{\scriptsize$\pm$0.00} & 78.8{\scriptsize$\pm$0.73} & \textbf{27.14{\scriptsize$\pm$0.03}} & \textbf{0.927{\scriptsize$\pm$0.00}} & \textbf{21.7{\scriptsize$\pm$0.21}} & 26.95{\scriptsize$\pm$0.04} & 0.920{\scriptsize$\pm$0.00} & 27.8{\scriptsize$\pm$0.24} \\
\addlinespace
\textbf{20} & 10.61{\scriptsize$\pm$0.01} & 0.482{\scriptsize$\pm$0.00} & 335.4{\scriptsize$\pm$0.81} & 18.98{\scriptsize$\pm$0.04} & 0.735{\scriptsize$\pm$0.00} & 160.6{\scriptsize$\pm$1.21} & \textbf{24.13{\scriptsize$\pm$0.03}} & \textbf{0.867{\scriptsize$\pm$0.00}} & \textbf{59.4{\scriptsize$\pm$0.44}} & \textbf{24.13{\scriptsize$\pm$0.03}} & 0.859{\scriptsize$\pm$0.00} & 75.3{\scriptsize$\pm$0.46} \\
\addlinespace
\textbf{30} & 8.39{\scriptsize$\pm$0.01} & 0.385{\scriptsize$\pm$0.00} & 422.6{\scriptsize$\pm$1.11} & 15.84{\scriptsize$\pm$0.05} & 0.644{\scriptsize$\pm$0.00} & 227.0{\scriptsize$\pm$1.37} & 22.04{\scriptsize$\pm$0.03} & \textbf{0.812{\scriptsize$\pm$0.00}} & \textbf{100.2{\scriptsize$\pm$0.58}} & \textbf{22.47{\scriptsize$\pm$0.03}} & 0.809{\scriptsize$\pm$0.00} & 116.5{\scriptsize$\pm$0.67} \\
\addlinespace
\textbf{40} & 7.18{\scriptsize$\pm$0.01} & 0.329{\scriptsize$\pm$0.00} & 427.0{\scriptsize$\pm$0.93} & 13.47{\scriptsize$\pm$0.05} & 0.582{\scriptsize$\pm$0.00} & 264.9{\scriptsize$\pm$1.21} & 19.78{\scriptsize$\pm$0.03} & 0.757{\scriptsize$\pm$0.00} & \textbf{137.0{\scriptsize$\pm$0.77}} & \textbf{21.25{\scriptsize$\pm$0.03}} & \textbf{0.768{\scriptsize$\pm$0.00}} & 141.1{\scriptsize$\pm$0.89} \\
\addlinespace
\textbf{50} & 6.59{\scriptsize$\pm$0.02} & 0.288{\scriptsize$\pm$0.00} & 402.3{\scriptsize$\pm$1.01} & 12.03{\scriptsize$\pm$0.05} & 0.531{\scriptsize$\pm$0.00} & 288.0{\scriptsize$\pm$1.16} & 16.87{\scriptsize$\pm$0.02} & 0.690{\scriptsize$\pm$0.00} & 179.1{\scriptsize$\pm$0.80} & \textbf{20.22{\scriptsize$\pm$0.03}} & \textbf{0.732{\scriptsize$\pm$0.00}} & \textbf{145.8{\scriptsize$\pm$0.78}} \\
\addlinespace
\textbf{60} & 6.29{\scriptsize$\pm$0.02} & 0.252{\scriptsize$\pm$0.00} & 415.6{\scriptsize$\pm$1.05} & 11.08{\scriptsize$\pm$0.05} & 0.485{\scriptsize$\pm$0.00} & 321.5{\scriptsize$\pm$1.11} & 13.84{\scriptsize$\pm$0.02} & 0.607{\scriptsize$\pm$0.00} & 233.2{\scriptsize$\pm$0.95} & \textbf{19.25{\scriptsize$\pm$0.03}} & \textbf{0.699{\scriptsize$\pm$0.00}} & \textbf{136.3{\scriptsize$\pm$0.77}} \\
\addlinespace
\textbf{70} & 6.06{\scriptsize$\pm$0.02} & 0.237{\scriptsize$\pm$0.00} & 468.7{\scriptsize$\pm$0.95} & 10.52{\scriptsize$\pm$0.04} & 0.447{\scriptsize$\pm$0.00} & 364.2{\scriptsize$\pm$1.08} & 11.53{\scriptsize$\pm$0.02} & 0.520{\scriptsize$\pm$0.00} & 283.8{\scriptsize$\pm$1.04} & \textbf{18.24{\scriptsize$\pm$0.03}} & \textbf{0.668{\scriptsize$\pm$0.00}} & \textbf{117.3{\scriptsize$\pm$0.65}} \\
\addlinespace
\textbf{80} & 5.91{\scriptsize$\pm$0.02} & 0.249{\scriptsize$\pm$0.00} & 514.5{\scriptsize$\pm$1.04} & 10.38{\scriptsize$\pm$0.04} & 0.418{\scriptsize$\pm$0.00} & 376.7{\scriptsize$\pm$1.20} & 10.24{\scriptsize$\pm$0.02} & 0.455{\scriptsize$\pm$0.00} & 302.3{\scriptsize$\pm$1.10} & \textbf{16.96{\scriptsize$\pm$0.03}} & \textbf{0.632{\scriptsize$\pm$0.00}} & \textbf{105.4{\scriptsize$\pm$0.60}} \\
\addlinespace
\textbf{90} & 5.74{\scriptsize$\pm$0.02} & 0.275{\scriptsize$\pm$0.00} & 468.2{\scriptsize$\pm$1.04} & 10.64{\scriptsize$\pm$0.03} & 0.388{\scriptsize$\pm$0.00} & 391.6{\scriptsize$\pm$1.15} & 9.57{\scriptsize$\pm$0.02} & 0.407{\scriptsize$\pm$0.00} & 284.1{\scriptsize$\pm$1.04} & \textbf{14.57{\scriptsize$\pm$0.03}} & \textbf{0.577{\scriptsize$\pm$0.00}} & \textbf{137.7{\scriptsize$\pm$0.70}} \\
\bottomrule
\end{tabular}}
\caption{\textbf{Image Inpainting:} CheXmix (S1 + S2) improves inpainting performance, with the masked autoregressive model showing notable advantages at higher masking percentages (Best metrics are in \textbf{bold}). We compute PSNR, MS-SSIM, and FID on a random sample of 5,000 images and report mean and standard deviation across three runs with different random seeds.}
\label{tab:image_quality_metrics_no_mae}
\end{table*}

\begin{table}[t]
\centering
\small
\setlength{\tabcolsep}{4pt}

% --- Green Score Table ---
\begin{tabular}{c|ccccc}
\multicolumn{6}{c}{\textbf{GREEN Score}} \\
\toprule
\textbf{Mask \%} & \textbf{Chameleon} & \textbf{RadPhi-2} & \textbf{CheXagent} & \textbf{CheXmix (S1)} & \textbf{CheXmix (S1 + S2)} \\
\midrule
\textbf{0}  & 0.019 {\tiny ±0.005} & 0.014 {\tiny ±0.004} & 0.152 {\tiny ±0.011} & 0.219 {\tiny ±0.015} & \textbf{0.221 {\tiny ±0.015}} \\
\textbf{10} & 0.017 {\tiny ±0.005} & 0.005 {\tiny ±0.002} & 0.146 {\tiny ±0.011} & 0.200 {\tiny ±0.015} & \textbf{0.220 {\tiny ±0.015}} \\
\textbf{20} & 0.039 {\tiny ±0.007} & 0.002 {\tiny ±0.001} & 0.134 {\tiny ±0.009} & 0.181 {\tiny ±0.013} & \textbf{0.215 {\tiny ±0.015}} \\
\textbf{30} & 0.012 {\tiny ±0.003} & 0.003 {\tiny ±0.002} & 0.133 {\tiny ±0.010} & 0.147 {\tiny ±0.012} & \textbf{0.217 {\tiny ±0.015}} \\
\textbf{40} & 0.022 {\tiny ±0.004} & 0.002 {\tiny ±0.001} & 0.117 {\tiny ±0.010} & 0.148 {\tiny ±0.012} & \textbf{0.224 {\tiny ±0.015}} \\
\textbf{50} & 0.015 {\tiny ±0.004} & 0.005 {\tiny ±0.002} & 0.112 {\tiny ±0.009} & 0.122 {\tiny ±0.012} & \textbf{0.215 {\tiny ±0.015}} \\
\textbf{60} & 0.017 {\tiny ±0.004} & 0.003 {\tiny ±0.002} & 0.103 {\tiny ±0.009} & 0.073 {\tiny ±0.009} & \textbf{0.217 {\tiny ±0.015}} \\
\textbf{70} & 0.019 {\tiny ±0.006} & 0.003 {\tiny ±0.002} & 0.085 {\tiny ±0.008} & 0.052 {\tiny ±0.008} & \textbf{0.192 {\tiny ±0.014}} \\
\textbf{80} & 0.015 {\tiny ±0.005} & 0.006 {\tiny ±0.002} & 0.068 {\tiny ±0.008} & 0.051 {\tiny ±0.006} & \textbf{0.176 {\tiny ±0.014}} \\
\textbf{90} & 0.010 {\tiny ±0.003} & 0.010 {\tiny ±0.003} & 0.052 {\tiny ±0.007} & 0.037 {\tiny ±0.008} & \textbf{0.094 {\tiny ±0.009}} \\
\bottomrule
\end{tabular}

\vspace{0.1em} 

% --- CheXbert-F1 Table ---
\begin{tabular}{c|ccccc}
\multicolumn{6}{c}{\textbf{CheXbert-F1}} \\
\toprule
\textbf{Mask \%} & \textbf{Chameleon} & \textbf{RadPhi-2} & \textbf{CheXagent} & \textbf{CheXmix (S1)} & \textbf{CheXmix (S1 + S2)} \\
\midrule
\textbf{0}  & 0.202 {\tiny ±0.018} & 0.154 {\tiny ±0.018} & 0.383 {\tiny ±0.022} & \textbf{0.390 {\tiny ±0.022}} & \textbf{0.390 {\tiny ±0.023}} \\
\textbf{10} & 0.164 {\tiny ±0.018} & 0.156 {\tiny ±0.017} & \textbf{0.403 {\tiny ±0.023}} & 0.396 {\tiny ±0.022} & 0.398 {\tiny ±0.023} \\
\textbf{20} & 0.178 {\tiny ±0.018} & 0.154 {\tiny ±0.019} & 0.357 {\tiny ±0.023} & \textbf{0.396 {\tiny ±0.022}} & \textbf{0.396 {\tiny ±0.023}} \\
\textbf{30} & 0.170 {\tiny ±0.019} & 0.131 {\tiny ±0.018} & 0.354 {\tiny ±0.023} & 0.389 {\tiny ±0.022} & \textbf{0.392 {\tiny ±0.022}} \\
\textbf{40} & 0.171 {\tiny ±0.019} & 0.140 {\tiny ±0.019} & 0.343 {\tiny ±0.022} & 0.376 {\tiny ±0.022} & \textbf{0.417 {\tiny ±0.023}} \\
\textbf{50} & 0.182 {\tiny ±0.019} & 0.115 {\tiny ±0.017} & 0.325 {\tiny ±0.021} & 0.380 {\tiny ±0.022} & \textbf{0.423 {\tiny ±0.023}} \\
\textbf{60} & 0.202 {\tiny ±0.020} & 0.129 {\tiny ±0.017} & 0.278 {\tiny ±0.021} & 0.359 {\tiny ±0.022} & \textbf{0.415 {\tiny ±0.023}} \\
\textbf{70} & 0.182 {\tiny ±0.019} & 0.098 {\tiny ±0.015} & 0.269 {\tiny ±0.020} & 0.374 {\tiny ±0.022} & \textbf{0.422 {\tiny ±0.023}} \\
\textbf{80} & 0.184 {\tiny ±0.019} & 0.100 {\tiny ±0.015} & 0.185 {\tiny ±0.017} & 0.343 {\tiny ±0.021} & \textbf{0.422 {\tiny ±0.023}} \\
\textbf{90} & 0.166 {\tiny ±0.019} & 0.122 {\tiny ±0.016} & 0.112 {\tiny ±0.015} & 0.283 {\tiny ±0.022} & \textbf{0.400 {\tiny ±0.023}} \\
\bottomrule
\end{tabular}

\vspace{0.1em} 

% --- RadGraph-F1 Table ---
\begin{tabular}{c|ccccc}
\multicolumn{6}{c}{\textbf{RadGraph-F1}} \\
\toprule
\textbf{Mask \%} & \textbf{Chameleon} & \textbf{RadPhi-2} & \textbf{CheXagent} & \textbf{CheXmix (S1)} & \textbf{CheXmix (S1 + S2)} \\
\midrule
\textbf{0}  & 0.026 {\tiny ±0.003} & 0.006 {\tiny ±0.002} & 0.091 {\tiny ±0.007} & \textbf{0.110 {\tiny ±0.008}} & 0.104 {\tiny ±0.008} \\
\textbf{10} & 0.022 {\tiny ±0.002} & 0.002 {\tiny ±0.001} & 0.090 {\tiny ±0.006} & 0.101 {\tiny ±0.008} & \textbf{0.106 {\tiny ±0.008}} \\
\textbf{20} & 0.032 {\tiny ±0.003} & 0.001 {\tiny ±0.001} & 0.083 {\tiny ±0.006} & 0.087 {\tiny ±0.007} & \textbf{0.102 {\tiny ±0.008}} \\
\textbf{30} & 0.017 {\tiny ±0.002} & 0.000 {\tiny ±0.000} & 0.073 {\tiny ±0.006} & 0.073 {\tiny ±0.006} & \textbf{0.102 {\tiny ±0.008}} \\
\textbf{40} & 0.013 {\tiny ±0.002} & 0.001 {\tiny ±0.001} & 0.068 {\tiny ±0.006} & 0.068 {\tiny ±0.005} & \textbf{0.109 {\tiny ±0.009}} \\
\textbf{50} & 0.016 {\tiny ±0.002} & 0.000 {\tiny ±0.000} & 0.062 {\tiny ±0.005} & 0.045 {\tiny ±0.004} & \textbf{0.102 {\tiny ±0.008}} \\
\textbf{60} & 0.014 {\tiny ±0.002} & 0.000 {\tiny ±0.000} & 0.048 {\tiny ±0.005} & 0.032 {\tiny ±0.004} & \textbf{0.107 {\tiny ±0.009}} \\
\textbf{70} & 0.014 {\tiny ±0.002} & 0.000 {\tiny ±0.000} & 0.045 {\tiny ±0.005} & 0.024 {\tiny ±0.003} & \textbf{0.098 {\tiny ±0.008}} \\
\textbf{80} & 0.014 {\tiny ±0.002} & 0.000 {\tiny ±0.000} & 0.033 {\tiny ±0.004} & 0.027 {\tiny ±0.003} & \textbf{0.077 {\tiny ±0.007}} \\
\textbf{90} & 0.009 {\tiny ±0.002} & 0.000 {\tiny ±0.000} & 0.030 {\tiny ±0.004} & 0.008 {\tiny ±0.003} & \textbf{0.042 {\tiny ±0.004}} \\
\bottomrule
\end{tabular}

\vspace{0.1em} 

% --- BERTScore Table ---
\begin{tabular}{c|ccccc}
\multicolumn{6}{c}{\textbf{BERTScore}} \\
\toprule
\textbf{Mask \%} & \textbf{Chameleon} & \textbf{RadPhi-2} & \textbf{CheXagent} & \textbf{CheXmix (S1)} & \textbf{CheXmix (S1 + S2)} \\
\midrule
\textbf{0}  & 0.243 {\tiny ±0.007} & 0.071 {\tiny ±0.008} & 0.320 {\tiny ±0.006} & 0.489 {\tiny ±0.008} & \textbf{0.503 {\tiny ±0.008}} \\
\textbf{10} & 0.241 {\tiny ±0.007} & 0.035 {\tiny ±0.009} & 0.318 {\tiny ±0.006} & 0.483 {\tiny ±0.009} & \textbf{0.505 {\tiny ±0.008}} \\
\textbf{20} & 0.253 {\tiny ±0.007} & 0.016 {\tiny ±0.006} & 0.310 {\tiny ±0.005} & 0.477 {\tiny ±0.008} & \textbf{0.499 {\tiny ±0.008}} \\
\textbf{30} & 0.213 {\tiny ±0.009} & 0.007 {\tiny ±0.006} & 0.306 {\tiny ±0.005} & 0.455 {\tiny ±0.009} & \textbf{0.506 {\tiny ±0.007}} \\
\textbf{40} & 0.161 {\tiny ±0.010} & 0.002 {\tiny ±0.011} & 0.305 {\tiny ±0.005} & 0.463 {\tiny ±0.007} & \textbf{0.508 {\tiny ±0.008}} \\
\textbf{50} & 0.175 {\tiny ±0.009} & 0.079 {\tiny ±0.017} & 0.295 {\tiny ±0.005} & 0.430 {\tiny ±0.007} & \textbf{0.500 {\tiny ±0.008}} \\
\textbf{60} & 0.155 {\tiny ±0.009} & 0.044 {\tiny ±0.005} & 0.291 {\tiny ±0.005} & 0.394 {\tiny ±0.009} & \textbf{0.505 {\tiny ±0.009}} \\
\textbf{70} & 0.154 {\tiny ±0.009} & 0.071 {\tiny ±0.005} & 0.287 {\tiny ±0.005} & 0.381 {\tiny ±0.009} & \textbf{0.501 {\tiny ±0.008}} \\
\textbf{80} & 0.164 {\tiny ±0.009} & 0.054 {\tiny ±0.006} & 0.271 {\tiny ±0.005} & 0.360 {\tiny ±0.008} & \textbf{0.480 {\tiny ±0.008}} \\
\textbf{90} & 0.131 {\tiny ±0.013} & 0.029 {\tiny ±0.006} & 0.263 {\tiny ±0.005} & 0.272 {\tiny ±0.009} & \textbf{0.450 {\tiny ±0.007}} \\
\bottomrule
\end{tabular}

\caption{\textbf{Radiology report generation.} CheXmix (S1 + S2) achieves the best performance across masking percentages (best metrics in \textbf{bold}). We compute GREEN score, CheXbert-F1, RadGraph-F1, and BERTScore on a random sample of 1,000 radiology reports and report mean and standard deviation across three runs with different random seeds.}
\label{tab:report_gen_masking_by_metric}
\end{table}
 \label{appendx:rrg_quantative_extended}
\pagestyle{empty}\begin{table}[h!]
\centering
\small
\setlength{\tabcolsep}{3.5pt}
\resizebox{0.48\textwidth}{!}{%
\begin{tabular}{c|cc|cc}
\toprule
\multirow{2}{*}{\textbf{Mask \%}} & \multicolumn{2}{c|}{\textbf{CheXmix S1}} & \multicolumn{2}{c}{\textbf{CheXmix S1 + S2}} \\
 & \textbf{B} (No CM) & \textbf{CM} (Causal) & \textbf{B} (No CM) & \textbf{CM} (Causal) \\
\midrule

% ---------------- CLASSIFICATION ----------------
\multicolumn{5}{l}{\cellcolor{gray!10}\textbf{a) CheXpert Classification}} \\
\multicolumn{5}{l}{\textit{AUROC ($\uparrow$)}} \\
0 & 0.713 {\tiny ±0.000} & 0.664 {\tiny ±0.000} & \textbf{0.716 {\tiny ±0.000}} & 0.712 {\tiny ±0.000} \\
20 & 0.702 {\tiny ±0.000} & 0.660 {\tiny ±0.000} & 0.684 {\tiny ±0.000} & \textbf{0.705 {\tiny ±0.000}} \\
40 & 0.678 {\tiny ±0.000} & 0.653 {\tiny ±0.001} & 0.691 {\tiny ±0.000} & \textbf{0.702 {\tiny ±0.000}} \\
60 & 0.667 {\tiny ±0.000} & 0.643 {\tiny ±0.000} & 0.682 {\tiny ±0.000} & \textbf{0.689 {\tiny ±0.000}} \\
80 & 0.591 {\tiny ±0.001} & 0.624 {\tiny ±0.000} & 0.632 {\tiny ±0.000} & \textbf{0.656 {\tiny ±0.000}} \\
\addlinespace[2pt]
\multicolumn{5}{l}{\textit{AUPRC ($\uparrow$)}} \\
0 & \textbf{0.382 {\tiny ±0.000}} & 0.333 {\tiny ±0.000} & 0.364 {\tiny ±0.000} & 0.377 {\tiny ±0.000} \\
20 & \textbf{0.364 {\tiny ±0.001}} & 0.339 {\tiny ±0.001} & 0.336 {\tiny ±0.001} & 0.344 {\tiny ±0.001} \\
40 & 0.340 {\tiny ±0.001} & 0.331 {\tiny ±0.001} & 0.345 {\tiny ±0.000} & \textbf{0.368 {\tiny ±0.001}} \\
60 & 0.325 {\tiny ±0.000} & 0.322 {\tiny ±0.000} & 0.343 {\tiny ±0.000} & \textbf{0.358 {\tiny ±0.002}} \\
80 & 0.277 {\tiny ±0.001} & 0.281 {\tiny ±0.001} & 0.301 {\tiny ±0.000} & \textbf{0.337 {\tiny ±0.001}} \\
\midrule

% ---------------- INPAINTING ----------------
\multicolumn{5}{l}{\cellcolor{gray!10}\textbf{b) Image Inpainting}} \\
\multicolumn{5}{l}{\textit{PSNR ($\uparrow$)}} \\
20 & 23.30 {\tiny ±0.01} & 24.13 {\tiny ±0.03} & \textbf{24.57 {\tiny ±0.03}} & 24.13 {\tiny ±0.03} \\
40 & 18.94 {\tiny ±0.01} & 19.78 {\tiny ±0.03} & \textbf{21.93 {\tiny ±0.03}} & 21.25 {\tiny ±0.03} \\
60 & 14.65 {\tiny ±0.01} & 13.84 {\tiny ±0.02} & \textbf{20.30 {\tiny ±0.03}} & 19.25 {\tiny ±0.03} \\
80 & 11.57 {\tiny ±0.01} & 10.24 {\tiny ±0.02} & \textbf{18.54 {\tiny ±0.03}} & 16.96 {\tiny ±0.03} \\
\addlinespace[2pt]
\multicolumn{5}{l}{\textit{MS-SSIM ($\uparrow$)}} \\
20 & 0.835 {\tiny ±0.00} & \textbf{0.867 {\tiny ±0.00}} & 0.863 {\tiny ±0.00} & 0.859 {\tiny ±0.00} \\
40 & 0.675 {\tiny ±0.00} & 0.757 {\tiny ±0.00} & \textbf{0.775 {\tiny ±0.00}} & 0.768 {\tiny ±0.00} \\
60 & 0.500 {\tiny ±0.00} & 0.607 {\tiny ±0.00} & \textbf{0.713 {\tiny ±0.00}} & 0.699 {\tiny ±0.00} \\
80 & 0.346 {\tiny ±0.00} & 0.455 {\tiny ±0.00} & \textbf{0.656 {\tiny ±0.00}} & 0.632 {\tiny ±0.00} \\
\addlinespace[2pt]
\multicolumn{5}{l}{\textit{FID ($\downarrow$)}} \\
20 & 107.1 {\tiny ±0.8} & \textbf{59.4 {\tiny ±0.4}} & 77.7 {\tiny ±0.5} & 75.3 {\tiny ±0.5} \\
40 & 251.3 {\tiny ±0.9} & \textbf{137.0 {\tiny ±0.8}} & 163.9 {\tiny ±0.8} & 141.1 {\tiny ±0.9} \\
60 & 314.5 {\tiny ±1.0} & 233.2 {\tiny ±1.0} & 154.6 {\tiny ±0.8} & \textbf{136.3 {\tiny ±0.8}} \\
80 & 332.3 {\tiny ±1.1} & 302.3 {\tiny ±1.1} & \textbf{69.50 {\tiny ±0.6}} & 105.4 {\tiny ±0.6} \\
\midrule

% ---------------- REPORT GEN ----------------
\multicolumn{5}{l}{\cellcolor{gray!10}\textbf{c) Report Generation}} \\
\multicolumn{5}{l}{\textit{GREEN Score ($\uparrow$)}} \\
0 & 0.172 {\tiny ±0.02} & 0.219 {\tiny ±0.02} & 0.140 {\tiny ±0.01} & \textbf{0.221 {\tiny ±0.02}} \\
20 & 0.153 {\tiny ±0.01} & 0.181 {\tiny ±0.01} & 0.141 {\tiny ±0.01} & \textbf{0.215 {\tiny ±0.01}} \\
40 & 0.112 {\tiny ±0.01} & 0.148 {\tiny ±0.01} & 0.132 {\tiny ±0.01} & \textbf{0.224 {\tiny ±0.01}} \\
60 & 0.071 {\tiny ±0.01} & 0.073 {\tiny ±0.01} & 0.129 {\tiny ±0.01} & \textbf{0.217 {\tiny ±0.01}} \\
80 & 0.061 {\tiny ±0.01} & 0.051 {\tiny ±0.01} & 0.104 {\tiny ±0.01} & \textbf{0.176 {\tiny ±0.01}} \\
\addlinespace[2pt]
\multicolumn{5}{l}{\textit{CheXbert F1 ($\uparrow$)}} \\
0 & 0.358 {\tiny ±0.02} & \textbf{0.390 {\tiny ±0.02}} & 0.300 {\tiny ±0.02} & \textbf{0.390 {\tiny ±0.02}} \\
20 & 0.336 {\tiny ±0.02} & \textbf{0.396 {\tiny ±0.02}} & 0.305 {\tiny ±0.02} & \textbf{0.396 {\tiny ±0.02}} \\
40 & 0.343 {\tiny ±0.02} & 0.376 {\tiny ±0.02} & 0.300 {\tiny ±0.02} & \textbf{0.417 {\tiny ±0.02}} \\
60 & 0.339 {\tiny ±0.02} & 0.359 {\tiny ±0.02} & 0.304 {\tiny ±0.02} & \textbf{0.415 {\tiny ±0.02}} \\
80 & 0.335 {\tiny ±0.02} & 0.343 {\tiny ±0.02} & 0.305 {\tiny ±0.02} & \textbf{0.422 {\tiny ±0.02}} \\
\addlinespace[2pt]
\multicolumn{5}{l}{\textit{RadGraph F1 ($\uparrow$)}} \\
0 & 0.086 {\tiny ±0.01} & \textbf{0.110 {\tiny ±0.01}} & 0.063 {\tiny ±0.01} & 0.104 {\tiny ±0.01} \\
20 & 0.071 {\tiny ±0.01} & 0.087 {\tiny ±0.01} & 0.062 {\tiny ±0.01} & \textbf{0.102 {\tiny ±0.01}} \\
40 & 0.054 {\tiny ±0.01} & 0.068 {\tiny ±0.01} & 0.062 {\tiny ±0.01} & \textbf{0.109 {\tiny ±0.01}} \\
60 & 0.024 {\tiny ±0.01} & 0.032 {\tiny ±0.00} & 0.059 {\tiny ±0.01} & \textbf{0.107 {\tiny ±0.01}} \\
80 & 0.012 {\tiny ±0.00} & 0.027 {\tiny ±0.00} & 0.040 {\tiny ±0.01} & \textbf{0.077 {\tiny ±0.01}} \\
\addlinespace[2pt]
\multicolumn{5}{l}{\textit{BERTScore ($\uparrow$)}} \\
0 & 0.483 {\tiny ±0.01} & 0.489 {\tiny ±0.01} & 0.437 {\tiny ±0.01} & \textbf{0.503 {\tiny ±0.01}} \\
20 & 0.468 {\tiny ±0.01} & 0.477 {\tiny ±0.01} & 0.439 {\tiny ±0.01} & \textbf{0.499 {\tiny ±0.01}} \\
40 & 0.448 {\tiny ±0.01} & 0.463 {\tiny ±0.01} & 0.432 {\tiny ±0.01} & \textbf{0.508 {\tiny ±0.01}} \\
60 & 0.458 {\tiny ±0.01} & 0.394 {\tiny ±0.01} & 0.430 {\tiny ±0.01} & \textbf{0.505 {\tiny ±0.01}} \\
80 & 0.454 {\tiny ±0.01} & 0.360 {\tiny ±0.01} & 0.400 {\tiny ±0.01} & \textbf{0.480 {\tiny ±0.01}} \\
\bottomrule
\end{tabular}
}
\caption{\textbf{Extended Causal Mask Ablation.} We evaluate CheXmix pretrained with 50\% masking using either bidirectional attention (B) or a causal mask (CM) across three tasks: (a) classification, (b) image inpainting, and (c) report generation. For both classification and report generation, CheXmix S1+S2 with CM consistently achieves the strongest performance across masking ratios. Although CheXmix S1+S2 (B) performs slightly better on image inpainting metrics, the differences between B and CM are small, typically only a few percentage points in PSNR and MS-SSIM.}
\label{tab:full_ablation}
\end{table}

\clearpage

\begin{table}[h]
    \centering
    \begin{tabular}{lcccc}
        \toprule
        \textbf{Model} & \textbf{GREEN}$\uparrow$ & \textbf{CheXbert}$\uparrow$ & \textbf{RadGraph-F1}$\uparrow$ & \textbf{BERTScore}$\uparrow$ \\
        \midrule
        Chameleon & 0.0287 \scriptsize{$\pm$0.001} & 0.139 \scriptsize{$\pm$0.017} & 0.033 \scriptsize{$\pm$0.003} & 0.272 \scriptsize{$\pm$0.005} \\
        HealthGPT & 0.216 \scriptsize{$\pm$0.017} & 0.239 \scriptsize{$\pm$0.021} & 0.075 \scriptsize{$\pm$0.004} & 0.398 \scriptsize{$\pm$0.005} \\
        CheXmix (S1 + S2) & \textbf{0.217 \scriptsize{$\pm$0.016}} & \textbf{0.413 \scriptsize{$\pm$0.027}} & \textbf{0.076 \scriptsize{$\pm$0.006}} & \textbf{0.426 \scriptsize{$\pm$0.007}} \\
        \bottomrule
    \end{tabular}
    \vspace{-0.5em}
    \caption{\textbf{ReXGradient-160K Report Generation (External Validation).} CheXmix (S1 + S2) outperforms early-fusion generative models on the ReXGradient-160K report generation task. The metrics are GREEN, CheXbert, RadGraph-F1, and BERTScore on a random sample of 500 radiology reports from the test set. There is no masking used for this experiment. We report the mean and standard deviation across three runs with different random seeds. CheXagent had metrics of GREEN: 0.135 \scriptsize{$\pm$0.011}, CheXbert: 0.310 \scriptsize{$\pm$0.024}, RadGraph: 0.068 \scriptsize{$\pm$[std]}, BERTScore: 0.398 \scriptsize{$\pm$0.005} on this task.}
\end{table}

% \FloatBarrier

\subsection{Extended Ablations} \label{appendix:section_c2}

\paragraph{Masking Ratio Ablation:} We conduct linear probe experiments on the CheXpert findings classification task using embeddings pretrained with four masking ratios (25\%, 50\%, 75\%, and 90\%) over 100K steps to evaluate how masking affects representation quality (Table~\ref{tab:masking_ablation}). We find that 50\% masking yields the highest AUROC and AUPRC, and therefore select this ratio for CheXmix (S1 + S2) pretraining. Higher masking ratios (75\% and 90\%) produce comparable classification performance, with differences within 0.01 AUROC. While general-domain studies have reported benefits from higher masking ratios~\cite{he2022masked}, prior work on chest X-ray autoencoders suggests that lower masking ratios can be more effective~\cite{xing2023self}.

\begin{table}[h!]
\centering
\begin{tabular}{c | c | c}
\text{Mask Pct (\%)} & \text{AUROC} & \text{AUPRC} \\ 
\hline
25 & 0.672 & 0.337 \\ 
50 & \textbf{0.676} & \textbf{0.341} \\
75 & 0.669 & 0.334 \\ 
90 & 0.641 & 0.327 
\end{tabular}
\vspace{-0.5em}
\caption{\textbf{Masking Ratio Ablation.} Effect of different masking percentages during CheXmix (S1 + S2) pretraining on CheXpert classification performance. AUROC and AUPRC are reported for each masking ratio, showing that 50\% masking yields the best discriminative performance. Consequently, we pretrain CheXmix (S1 + S2) with 50\% masking.}
\label{tab:masking_ablation}
\end{table}

\paragraph{CheXmix S1 Extended:} CheXmix (S1) refers to an intermediate checkpoint obtained after Stage 1 training, without proceeding to Stage 2. On CheXpert classification, CheXmix S1-extended trained for additional S1 steps achieves similar AUROC performance compared to the original S1 checkpoint (0.667$\pm$0.001 vs. 0.664$\pm$0.000). Notably, despite having the same total number of training steps, CheXmix (S1 + S2) outperforms CheXmix S1-extended by 6.75\%, highlighting that Stage 2 multimodal masked training improves representation quality beyond simply increasing training duration.

\begin{table}[h]
\centering
\label{tab:chexmix_extended}
\resizebox{0.6\columnwidth}{!}{% % Use \columnwidth for 2-column or \textwidth for 1-column
\begin{tabular}{@{}lcc@{}}
\toprule
\textbf{CheXmix Variant} & \textbf{Total Training Steps} & \textbf{AUROC} \\ \midrule
CheXmix S1             & 703,671                       & $0.664 \pm 0.000$ \\
CheXmix S1-extended    & 1,217,664                     & $0.667 \pm 0.001$ \\
CheXmix (S1 + S2)             & 1,217,664                     & $\mathbf{0.712 \pm 0.001}$ \\ \bottomrule
\end{tabular}
}
\vspace{-0.5em}
\caption{\textbf{S1 Extended Ablation (no masking).} Comparison of CheXmix variants showing that additional Stage 1 training (S1-extended) yields minimal gains, while Stage 2 multimodal masked training strategy (S1 + S2) improves AUROC at the same total training budget.}
\end{table}

\clearpage

\FloatBarrier

\section{Extended Qualitative Results}
\pagestyle{plain}
\subsection{Image Inpainting Examples}

\begin{figure*}[ht]
% \thisfloatpagestyle{empty}
\centerline{\includegraphics[width=1.0\textwidth]{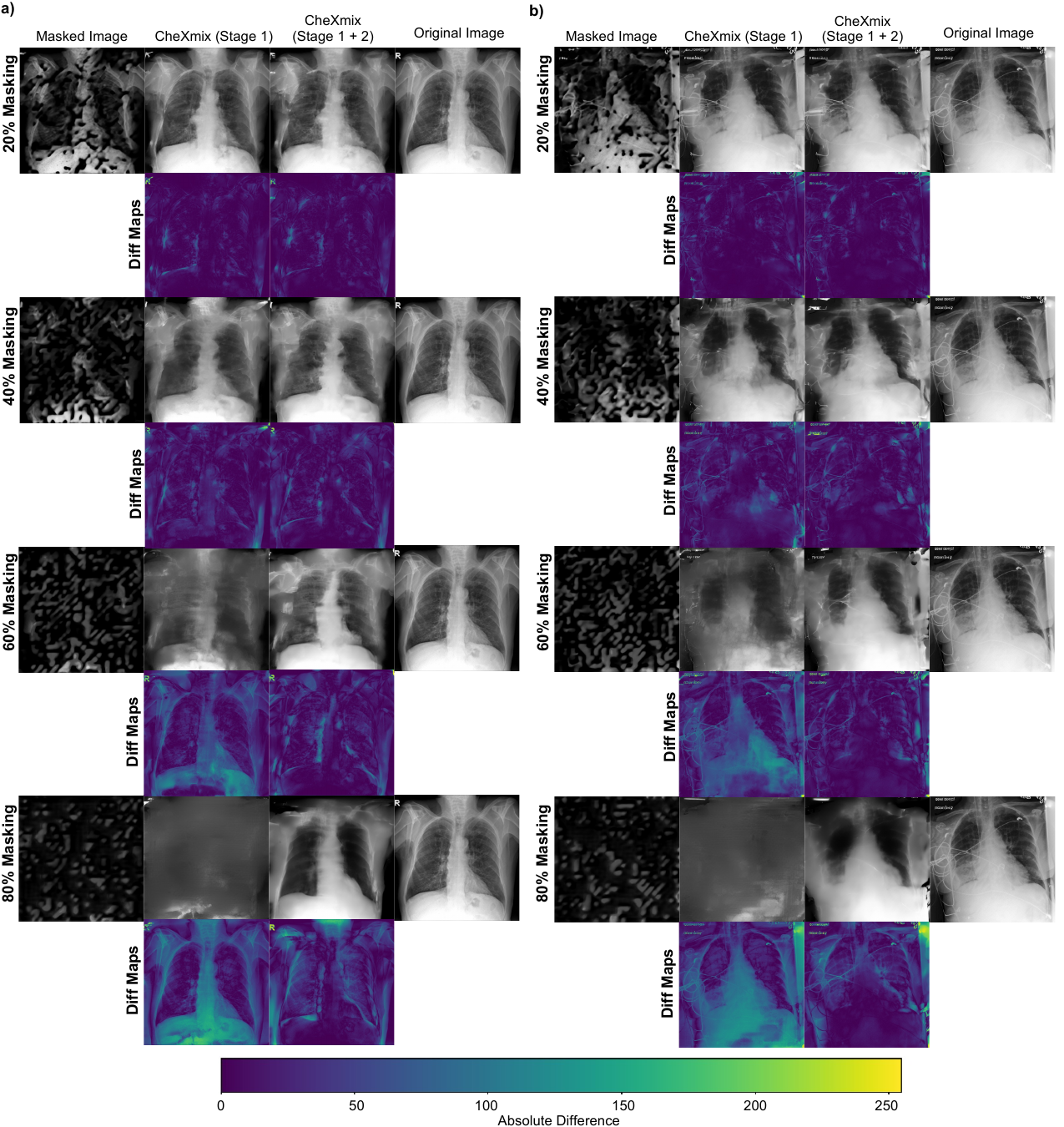}}
\caption{\textbf{Image Inpainting Visualization} CheXmix (S1 + S2) pretraining provides substantial improvements in inpainting quality at higher masking ratios for (a) a chest radiograph with COPD and (b) a case with consolidation.}
\label{fig:test_time_ensembling}
\end{figure*}

\clearpage
\subsection{Radiology Report Examples} \label{appendix:section_d2}
\label{appendix:rrg_qualitative}

\begin{table*}[h!]
\centering
\resizebox{0.9\textwidth}{!}{
\footnotesize % Use small font to fit the text
\renewcommand{\arraystretch}{1.3} % Add breathing room between rows
\setlength{\tabcolsep}{6pt}
\begin{tabularx}{\textwidth}{p{0.15\textwidth} X}
\toprule
\textbf{Model} & \textbf{Generated Report} \\

% ================= SAMPLE 1 =================
\midrule
\multicolumn{2}{c}{\cellcolor{gray!20}\textbf{Sample 1: Normal Case}} \\
\midrule
\textbf{Reference} & \textit{Findings:} The lungs are well inflated and clear. Heart size and mediastinal contours are normal. There is no pleural effusion or pneumothorax. Osseous structures are intact. \newline \textit{Impression:} No acute cardiopulmonary process. \\
\midrule
Chameleon & This is a posteroanterior chest radiograph, which is a type of X-ray image that is taken from the back and shows the front of the chest. The image appears to be of a patient with a medical condition, as evidenced by the presence of a tube or catheter in the right upper lobe of the lung, which is likely being used to treat a respiratory issue. The tube is attached to a device that is emitting a white, glowing light, which suggests that the tube is being used for some sort of medical procedure. The patient's heart is also visible on the X-ray, and it appears to be in good condition. The lungs are not fully inflated, which may indicate that the patient is experiencing some difficulty breathing or that they have recently been intubated. The ribs are also visible on the image, and they appear to be in good condition. The spine is also visible, and it appears to be straight and well-aligned. Overall, the image appears to show a patient with a medical condition that is being treated with a tube or catheter in the right upper lobe of the lung. \\
\midrule
CheXagent & [Cardiac: Heart Size] No acute cardiopulmonary process. \\
\midrule
CheXmix (S1) & \textit{Findings:} The lungs are clear without focal consolidation. No pleural effusion or pneumothorax is seen. The cardiac and mediastinal silhouettes are stable. \newline \textit{Impression:} No acute cardiopulmonary process. No significant interval change. No new focal \\
\midrule
\textbf{CheXmix (S1 + S2)} & \textit{Findings:} The lungs are clear without focal consolidation. No pleural effusion or pneumothorax is seen. The cardiac and mediastinal silhouettes are unremarkable. \newline \textit{Impression:} No acute cardiopulmonary process. No focal consolidation to suggest pneumonia. \\

% ================= SAMPLE 2 =================
\midrule
\multicolumn{2}{c}{\cellcolor{gray!20}\textbf{Sample 2: Case with pleural effusions and support devices}} \\
\midrule
\textbf{Reference} & \textit{Findings:} In comparison with the study of \_\_\_, the left chest tube and subclavian catheters remain in place. No evidence of pneumothorax. Cardiac silhouette is at the upper limits of normal in size. There is evidence of elevated pulmonary venous pressure. Mild haziness of the lower zones is consistent with small bilateral pleural effusions and compressive atelectasis at the bases. \textit{Impression:} \\
\midrule
Chameleon & The image depicts a chest x-ray of a patient with a suspected pulmonary embolism (PE). The findings on the x-ray include:
\begin{itemize}
    \item Right lower lobe consolidation with air bronchograms, suggesting an acute pulmonary embolism (PE)
    \item A large pleural effusion present on the right side, which could be related to the PE or a separate condition
    \item A small pleural effusion present on the left side, which may be related to the PE or a separate condition
    \item Cardiomegaly, which is suggestive of heart failure or other cardiac conditions
\end{itemize}

The impression based on the findings is that the patient likely has a pulmonary embolism (PE), as well as a possible cardiomyopathy or heart failure. The large pleural effusions on both sides may be related to the PE or a separate condition. Further testing, such as a CT pulmonary angiogram (CTPA) or ventilation-perfusion (VQ) scan, may be necessary to confirm the diagnosis and guide treatment. \\
\midrule
CheXagent & [Breathing: Lungs] Mild pulmonary edema is present. [Breathing: Pleura] A small right pleural effusion is noted. [Breathing: Lungs] Mild bibasilar atelectasis is observed. \\
\midrule
CheXmix (S1) & \textit{Findings: Impression:} AP chest compared to \_\_\_: Previous mild pulmonary edema has improved, but there is still substantial bibasilar atelectasis, left greater than right, and small pleural effusionss, left greater than right. Heart size is normal. ET tube and right internal jugular line are in standard positions. \\
\midrule
\textbf{CheXmix (S1 + S2)} & \textit{Findings: Impression:} In comparison with the study of \_\_\_, there is little change in the appearance of the heart and lungs. Again there is enlargement of the cardiac silhouette with pulmonary vascular congestion and bilateral pleural effusions with compressive atelectasis at the bases. The right IJ catheter extends to the mid portion of the SVC. No evidence of pneumothorax. \\
\bottomrule
\end{tabularx}
}
\caption{\textbf{Qualitative comparison of generated reports.} Sample 1 demonstrates a normal case. Sample 2 demonstrates a complex case with support devices and pleural effusions. Generated reports from CheXmix (S1+S2) are shown for comparison.}
\label{tab:qualitative_results}
\end{table*}

\clearpage
\section{Test-Time Augmentation Prompt} 
\label{suppl_sec:TTA_prompt}

\begin{tcolorbox}[colback=gray!5, colframe=gray!50, sharp corners]
\ttfamily\small
You are an expert medical report synthesizer. Your task is to analyze five generated radiology reports below. These reports are variations describing the same patient.

\vspace{0.5em}
Compile them into one, single, consolidated report that contains:
\begin{enumerate}[nosep, leftmargin=*, label=\arabic*.]
    \item A single ``Findings'' section.
    \item A single ``Impressions'' section.
\end{enumerate}

\vspace{0.5em}
Rules:
\begin{enumerate}[nosep, leftmargin=*, label=\arabic*.]
    \item Combine and de-duplicate all repetitive information.
    \item Make sure the synthesized report is the SAME LENGTH as the original reports.
    \item If there are slight variations in wording for the same finding, use the most precise and complete description.
    \item Ensure the final ``Findings'' and ``Impressions'' are comprehensive and written as a single, coherent section each with no newlines or bullet points.
    \item If no findings or impressions are present in the generated reports, then leave blank.
\end{enumerate}
\end{tcolorbox}
% {
%     \small
%     \bibliographystyle{ieeenat_fullname}
%     \bibliography{main}
% }

\end{document}